\documentclass[10pt,twocolumn,letterpaper]{article}

\usepackage{cvpr}              
\usepackage[accsupp]{axessibility}

\usepackage[dvipsnames]{xcolor}

\definecolor{cvprblue}{rgb}{0.21,0.49,0.74}
\usepackage[pagebackref,breaklinks,colorlinks,citecolor=cvprblue]{hyperref}
\usepackage{amsmath}
\def\paperID{5929} 
\def\confName{CVPR}
\def\confYear{2024}

\title{Fantastic Animals and Where to Find Them: Segment Any Marine Animal with Dual SAM}

\author{Pingping Zhang\thanks{Corresponding author} \quad   Tianyu Yan  \quad  Yang Liu   \quad Huchuan Lu\\
School of Future Technology, School of Artificial Intelligence, Dalian University of Technology, China\\
{\tt\small 2981431354@mail.dlut.edu.cn; \{zhpp,ly,lhchuan\}@dlut.edu.cn}\\
}

\begin{document}
\maketitle
\begin{abstract}
As an important pillar of underwater intelligence, Marine Animal Segmentation (MAS) involves segmenting animals within marine environments.
Previous methods don't excel in extracting long-range contextual features and overlook the connectivity between discrete pixels.
Recently, Segment Anything Model (SAM) offers a universal framework for general segmentation tasks.
Unfortunately, trained with natural images, SAM does not obtain the prior knowledge from marine images.
In addition, the single-position prompt of SAM is very insufficient for prior guidance.
To address these issues, we propose a novel feature learning framework, named Dual-SAM for high-performance MAS.
To this end, we first introduce a dual structure with SAM's paradigm to enhance feature learning of marine images.
Then, we propose a Multi-level Coupled Prompt (MCP) strategy to instruct comprehensive underwater prior information, and enhance the multi-level features of SAM's encoder with adapters.
Subsequently, we design a Dilated Fusion Attention Module (DFAM) to progressively integrate multi-level features from SAM's encoder.
Finally, instead of directly predicting the masks of marine animals, we propose a Criss-Cross Connectivity Prediction (C$^3$P) paradigm to capture the inter-connectivity between discrete pixels.
With dual decoders, it generates pseudo-labels and achieves mutual supervision for complementary feature representations, resulting in considerable improvements over previous techniques.
Extensive experiments verify that our proposed method achieves state-of-the-art performances on five widely-used MAS datasets.
The code is available at \href{https://github.com/Drchip61/Dual_SAM}{https://github.com/Drchip61/Dual\_SAM}.
\end{abstract} 
\section{Introduction}
\label{sec:intro}
\begin{figure}[!t]
\centering
\hspace{-4mm}
\includegraphics[width=3.4in,height=7.8cm]{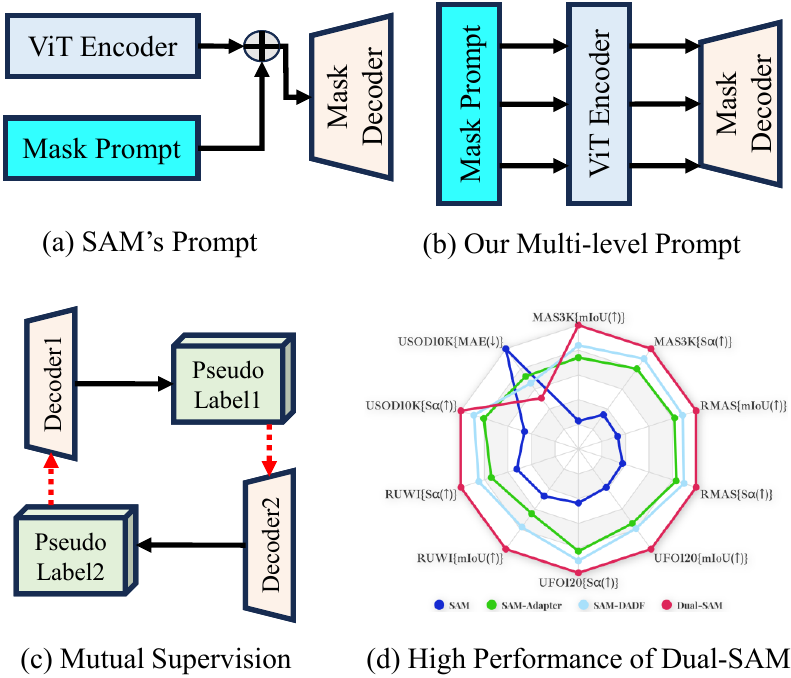}
\vspace{-6mm}
\caption{Our inspirations and advantages. (a) Single-position prompt of SAM. (b) Our multi-level prompt. (c) Mutual supervision for our Dual-SAM's decoders. (d) Our Dual-SAM delivers high performances on multiple datasets.}
\label{motivations}
\vspace{-6mm}
\end{figure}
Underwater ecosystems contain a wide variety of marine life, from microscopic plankton to colossal whales.
These ecosystems are crucial roles for the earth's environmental balance.
Accurate and efficient Marine Animal Segmentation (MAS) is vital for understanding species' distributions, behaviors, and interactions within the submerged world.
However, unlike conventional terrestrial images, underwater images include variable lighting conditions, water turbidity, color distortion, and the movement of both cameras and subjects.
Traditional segmentation techniques, developed primarily for terrestrial settings, often fall short when applied to the underwater domain.
Consequently, methods designed to tackle the unique aspects of the marine environment are urgently required for underwater intelligence.

With the advent of deep learning, Convolutional Neural Networks (CNNs)~\cite{he2016deep,huang2017densely} lead to a new era for image segmentation.
In fact, CNNs demonstrate a remarkable ability to extract intricate features, which makes them suitable for marine animal segmentation.
%
%
Nonetheless, CNNs have inherent limitations in capturing long-range dependencies and contextual information within an image.
Recently, Transformers~\cite{dosovitskiy2020image} offer enhanced performance in capturing the long-range features of complex images.
This ability is particularly appealing for underwater image segmentation, where the contextual information is often crucial to discern a marine organism from its background.
However, one significant challenge for Transformers is the need of vast amounts of training data.
Building on this evolution, the Segment Anything Model (SAM)~\cite{kirillov2023segment} utilizes one billion natural images for model training.
However, since the pre-training of SAM is primarily conducted under natural lighting conditions, its performance in marine environments is not optimal.
In addition, the simplicity of SAM's decoder limits its ability to capture complex details of marine organisms.
Moreover, SAM introduces external prompts for instructing object priors.
However, the single-position prompt is very insufficient for prior guidance.
%

To overcome the aforementioned issues, in this work we propose a novel feature learning framework, named Dual-SAM for high-performance MAS.
Fig.~\ref{motivations} shows our inspirations and advantages.
Technically, we first introduce a dual structure with SAM's paradigm to enhance feature learning of marine images with gamma correction operations.
Meanwhile, we enhance the multi-level features of SAM's encoder with adapters.
Then, we propose a Multi-level Coupled Prompt (MCP) strategy to instruct comprehensive underwater prior information with auto-prompts.
Subsequently, we design a Dilated Fusion Attention Module (DFAM) to progressively integrate multi-level features from SAM's encoder.
Finally, instead of directly predicting the masks of marine animals, we propose a Criss-Cross Connectivity Prediction (C$^3$P) paradigm to capture the inter-connectivity between discrete pixels.
With dual decoders, it generates pseudo-labels and achieves mutual supervision for complementary feature representations.
The proposed vectorized representation delivers significant improvements over previous scalar prediction techniques.
Extensive experiments show that our proposed method achieves state-of-the-art performances on five widely-used MAS datasets.

In summary, our contributions are listed as follows:
\begin{itemize}
\item We propose a novel feature learning framework, named Dual-SAM for Marine Animal Segmentation (MAS).
The framework inherits the ability of SAM and adaptively incorporates prior knowledge of underwater scenarios.
\item We propose a Multi-level Coupled Prompt (MCP) strategy to instruct comprehensive underwater prior information with auto-prompts.
\item We propose a Dilated Fusion Attention Module (DFAM) and a Criss-Cross Connectivity Prediction (C$^3$P) to improve the localization perception of marine animals.
\item We perform extensive experiments to verify the effectiveness of the proposed modules. Our approach achieves a new state-of-the-art performance on five MAS datasets.
\end{itemize}
\vspace{-4mm} 
\section{Related Work}
\label{sec:formatting}
\subsection{Marine Animal Segmentation}
MAS suffers from great challenges, such as variable lighting, particulate matter, water turbidity, etc.
%
%
In past decades, most of existing methods primarily utilize handcrafted features~\cite{ng2003sift,bay2008speeded,priyadharsini2019object}.
Technically, energy-based models~\cite{shihavuddin2013automated,lane1998robust,priyadarshni2020underwater} are usually employed to predict the binary masks of marine animals.
Although they achieve great success, there are still some key limitations, such as low robustness to the blurriness, unclear boundaries, etc.

With the rise of deep learning, CNNs become the preferred models for MAS.
Various network architectures have been proposed to achieve performance improvements.
For example, Li~\textit{et al.}~\cite{li2021marine} propose a feature-interactive encoder and a cascade decoder to extract more comprehensive information.
Liu \textit{et al.}~\cite{liu2022underwater} incorporate channel and spatial attention modules to refine the feature map for better object boundaries.
Furthermore, Chen \textit{et al.}~\cite{chen2022robust} extract multi-scale features and introduce attention fusion blocks to highlight marine animals.
Fu \textit{et al.}~\cite{fu2023masnet} design a data augmentation strategy and use a Siamese structure to learn shared semantic information.
Although effective, these CNN-based models lack the ability to capture long-range dependencies and intricate details for complex marine images.

Recently, Vision Transformer (ViT)~\cite{dosovitskiy2020image} presents an excellent global understanding ability for multiple data types.
With structural modifications, it delivers remarkable performances in various segmentation tasks~\cite{zheng2021rethinking,ranftl2021vision,yan2022fully,10292873}.
As for MAS, Hong \textit{et al.}~\cite{hong2023usod10k} adapt Transformer-based encoders to underwater images and show promising animal segmentation results.
However, one significant challenge for Transformers is the need of vast amounts of training data.
Currently, there are no very large-scale MAS datasets for the training of Transformers.
%
%
\subsection{Segment Anything Model for Customized Tasks}
Recently, SAM~\cite{kirillov2023segment} is proposed to achieve universal image segmentation.
It is trained on a large-scale segmentation dataset and exhibits zero-shot transfer capabilities~\cite{zhang2023segment,lei2023medlsam,zhang2023sam3d}.
With various types of prompts, it is efficiently deployed for a multitude of applications~\cite{zhao2023enlighten,jin2023let,shan2023robustness}.
However, it exhibits performance limitations in transfer scenarios.
In addition, the simplicity of SAM's decoder is a hindrance when dealing with detail-aware segmentation tasks.
%
\begin{figure*}[htb]
\centering
\includegraphics[width=\textwidth]{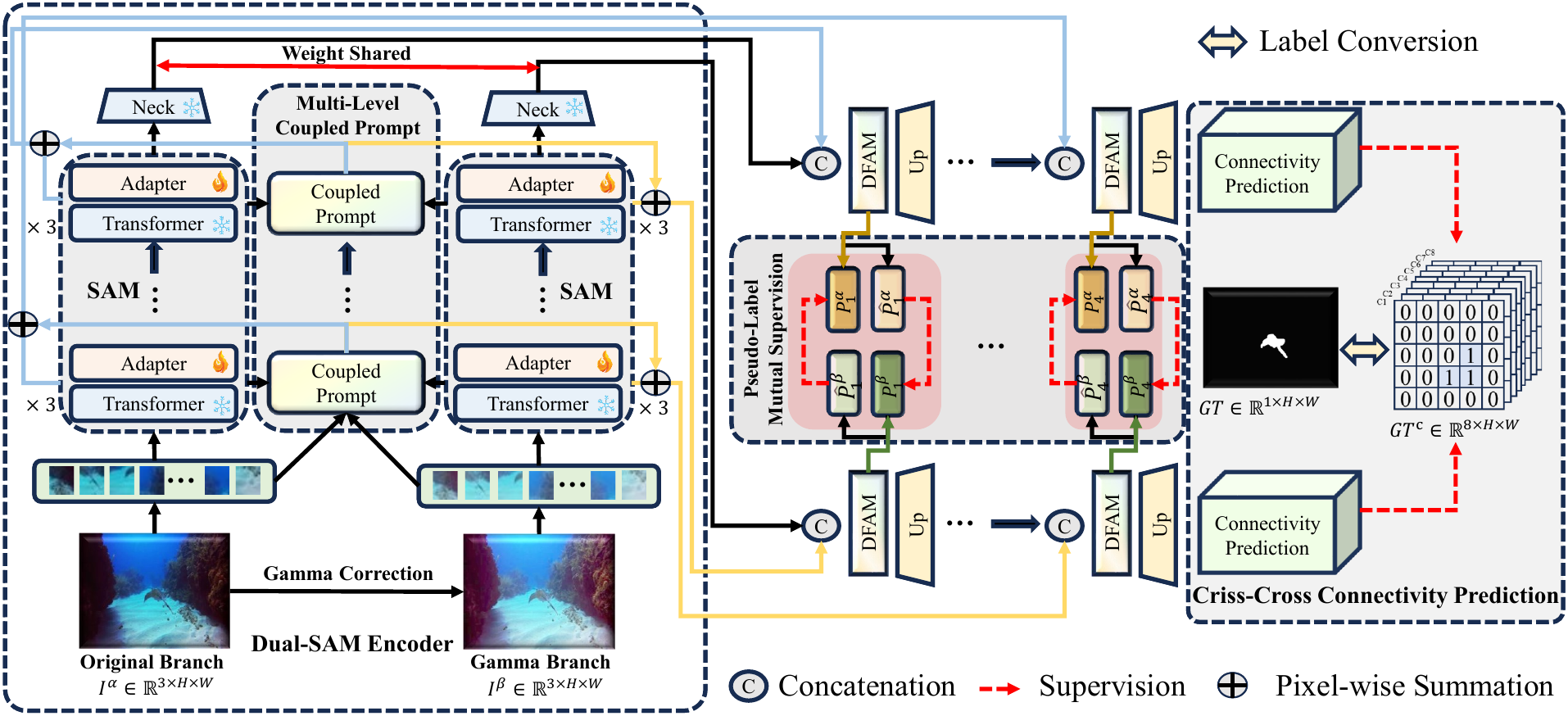}
\vspace{-4mm}
\caption{The whole framework of our proposed approach. It contains five main components: Dual-SAM Encoder (DSE), Multi-level Coupled Prompt (MCP), Dilated Fusion Attention Module (DFAM), Criss-Cross Connectivity Prediction (C$^3$P) and Pseudo-label Mutual Supervision (PMS). Our framework can significantly improve the Marine Animal Segmentation (MAS) with SAM.}
\vspace{-4mm}
\label{fig:framework}
\end{figure*}

To address these limitations, various approaches have been proposed.
Some works adopt adapters~\cite{zhang2023customized,chen2023sam,lai2023detect} to infuse SAM with domain-specific information.
Others have opted for more specific decoder structures~\cite{gao2023desam} to improve the domain perception.
There are also efforts to automate the generation of prompts~\cite{chen2023rsprompter} for a better adaptability.
Despite these advancements, since trained with natural images, SAM does not obtain enough prior knowledge from specific domains.
In addition, the single-position prompt of SAM is very insufficient for prior guidance.
%
%
As for MAS, we find that there is only one work~\cite{xu2023aquasam} involving fine-tuning SAM for underwater scenes.
Therefore, in this work, we delve deeply into SAM for improving the customized tasks.
%
%
\section{Proposed Approach}
As shown in Fig.~\ref{fig:framework}, our method contains five main components: Dual-SAM Encoder (DSE), Multi-level Coupled Prompt (MCP), Dilated Fusion Attention Module (DFAM), Criss-Cross Connectivity Prediction (C$^3$P) and Pseudo-label Mutual Supervision (PMS).
These components will be elaborated in the following subsections.
\subsection{Dual-SAM Encoder}
As previously mentioned, it is imperative to enhance marine images with characteristics of natural images.
To this end, we utilize the gamma correction for illumination compensation.
Given the marine image $I^{\alpha}$, the corrected image $I^{\beta}$ can be expressed as:
\begin{equation}
  I^{\beta}=\sqrt[\gamma]{I^{\alpha}}, \gamma=\text{lg}(0.5)-\text{lg}(mean_I^{gray}/255),
  \label{eq:important}
\end{equation}
where $\gamma$ is the gamma coefficient and ${mean}_I^{gray}$ is the mean value of the image's gray-scale intensities.

Afterwards, we inject marine domain information into SAM's encoder for a better marine feature extraction.
Inspired by~\cite{zhang2023customized, chen2023sam}, we employ low-rank trainable matrices~\cite{hu2021lora} to the Query and Value portion of the Multi-Head Self-Attention (MHSA) block.
In addition, we incorporate an Adapter~\cite{houlsby2019parameter} to the Feed-Forward Network (FFN).
Without loss of generality, let $X_j \in \mathbb{R}^{N \times D}$ be the output feature in the $j$-th layer of SAM's encoder, the feature in the $j+1$-th layer can be represented as follows:
\vspace{-2mm}
\begin{equation}
Q_{j}=X_{j}W_{q}+(X_{j}W_{q}^{\text {down}})W_{q}^{up},
\end{equation}
\vspace{-2mm}
\begin{equation}
K_{j}=X_{j}W_{k},
\end{equation}
\vspace{-2mm}
\begin{equation}
V_{j}=X_{j}W_{v}+(X_{j}W_{v}^{\text{down}})W_{v}^{up},
\end{equation}
\vspace{-2mm}
\begin{equation}
H_{j}=\text{MHSA}\left(Q_j,K_j,V_j\right)+X_{j},
\end{equation}
\vspace{-2mm}
\begin{equation}
X_{j+1}=\psi\left(\text{FFN}\left(\phi\left(H_{j}\right)\right) W^{down}\right) W^{up}+H_{j},
\end{equation}
where $N$ is the total number of tokens. $D$ is the dimension of the token embedding.
$W_{q/v}^{down}\in \mathbb{R}^{D \times r}$ and $W_{q/v}^{up}\in \mathbb{R}^{r \times D}$ are linear projection matrices that reduce and subsequently restore the dimension of features, respectively.
$r$ stands for the dimension to which the features are reduced.
$H_i$ is the intermediate features within the Transformer block.
Similarly, $W^{down} \in \mathbb{R}^{D \times R} $ and $W^{up} \in \mathbb{R}^{R \times D} $ are the compressed and excited operation, respectively.
$R$ stands for the compressed dimension.
$\psi$ is the GELU~\cite{hendrycks2016gaussian} activation function.
$\phi$ is the layer normalization.
%
%
Since we only update the linear projection matrices, it significantly reduces the number of trainable parameters for subsequent MAS tasks.
With an additional branch, it can enhance animal-related features for better localizing.
\begin{figure}[]
    \centering
    \resizebox{0.42\textwidth}{!}
	{
    \includegraphics[width=0.5\textwidth]{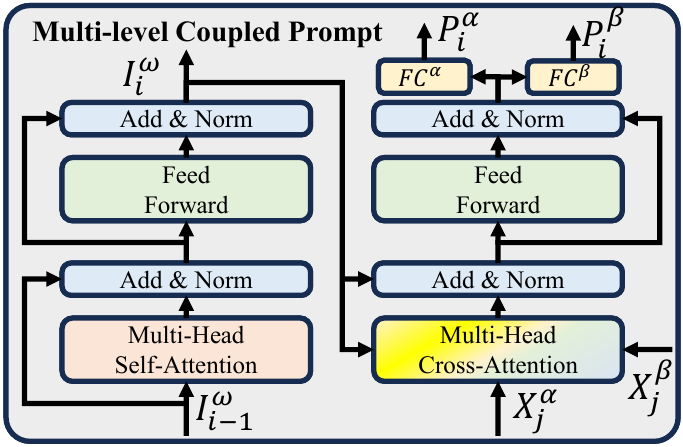}
    }
    \vspace{-2mm}
    \caption{Our proposed Multi-level Coupled Prompt (MCP).}
    \vspace{-4mm}
    \label{fig:crossprompt}
\end{figure}
\subsection{Multi-level Coupled Prompt}
In SAM, object-related prompts (e.g., mask, box, point) are encoded and added to the feature maps.
However, the single-position prompt is very insufficient for prior guidance.
%
%
To improve the prompt ability, we propose a Multi-level Coupled Prompt (MCP) strategy to instruct comprehensive underwater prior information with auto-prompts.

To this end, we first concatenate the original image $I^{\alpha}$ and the corrected image $I^{\beta}$.
Then, we partition them into patches and use convolutions to obtain feature embeddings:
\begin{equation}
I_{0}^\omega=\text{PatchEmbed}([I^\alpha,I^\beta]),
\label{eq:important}
\end{equation}
where $I_{0}^\omega \in \mathbb{R}^{N \times D}$ is the tokenized features, which can be served as the start point.
As shown in Fig.~\ref{fig:crossprompt}, it undergoes several Transformer layers and iteratively generate features:
\begin{equation}
I_{i}^{\omega}=\text{Trans}(I_{i-1}^{\omega}), i=1,2,3,4.
\end{equation}
Then, we treat the DSE's output features $X_{j}^{\alpha}$ and $X_{j}^{\beta}$ as the Query and Key, respectively.
By using $I_{i}^{\omega}$ as Value, we can obtain the coupled prompts as follows:
\begin{equation}
H_{i}^{\tau}=\text{MHCA}\left(X_{j}^{\alpha}, X_{j}^{\beta}, I_{i}^{\omega}\right)+I_{i}^{\omega},
\end{equation}
\vspace{-2mm}
\begin{equation}
\mathcal{P}_{i}^{\omega}=\text{FFN}\left(\phi\left(H_i^\tau\right)\right)+H_i^\tau,
\end{equation}
\vspace{-2mm}
\begin{equation}
\mathcal{P}_{i}^{\alpha}=\text{FC}^{\alpha}(\mathcal{P}_{i}^{\omega}),
\end{equation}
\vspace{-2mm}
\begin{equation}
\mathcal{P}_{i}^{\beta}=\text{FC}^{\beta}(\mathcal{P}_{i}^{\omega}),
\end{equation}
where MHCA is the Multi-Head Cross-Attention block and FC is a fully-connected layer.
The generated prompts ($\mathcal{P}_{i}^{\beta}$ and $\mathcal{P}_{i}^{\beta}$) are coupled and can be used as auto-prompts for a better instruction and prior guidance.
As a result, we can obtain prompted features by:
\begin{equation}
E_{i}^{\alpha}=X_{j}^{\alpha}+g_i^\alpha\mathcal{P}_{i}^{\alpha},
\label{eq:important}
\end{equation}
\vspace{-2mm}
\begin{equation}
E_{i}^{\beta}=X_{j}^{\beta}+g_i^\beta\mathcal{P}_{i}^{\beta},
\label{eq:also-important}
\end{equation}
where $g_i^\alpha$ and $g_i^\beta$ are learnable weights for balancing the input features and prompts.
%
\begin{figure}[]
\centering
\resizebox{0.5\textwidth}{!}
{
\includegraphics[width=0.5\textwidth]{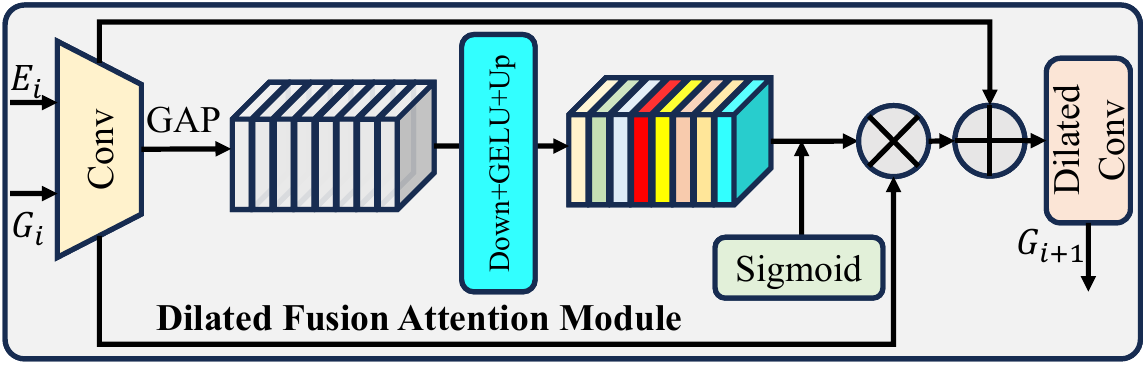}
}
\vspace{-6mm}
\caption{Our Dilated Fusion Attention Module (DFAM).}
\vspace{-4mm}
\label{fig:dfam}
\end{figure}
\subsection{Dilated Fusion Attention Module}
The simple decoder of SAM is a hindrance when dealing with complex segmentation tasks.
Inspired by~\cite{lin2017feature}, we introduce feature pyramid structures as decoders to fuse the prompted features for MAS.
To improve the receptive field, we propose the Dilated Fusion Attention Module (DFAM) with dilated convolution~\cite{chen2017deeplab} and channel attention.
It can be inserted in adjacent features ($G_{i}$ and $G_{i+1}$).
As shown in Fig.~\ref{fig:dfam}, the DFAM can be represented as follows:
\begin{equation}
F_{i}^{r}=\psi\left(\Theta_{1\times1}\left([E_{i}, G_{i}]\right)\right),
\end{equation}
\vspace{-2mm}
\begin{equation}
W^{g}=\sigma\left(\psi\left(\text{GAP}\left(F_{i}^{r}\right) W^{down}\right) W^{up}\right),
\end{equation}
\vspace{-2mm}
\begin{equation}
F_{i}=W^{g}F_{i}^{r}+F_{i}^{r},
\end{equation}
\vspace{-2mm}
\begin{equation}
G_{i+1}=\psi\left(\Theta_{3,3}^{2}\left(F_{i}\right)\right),
\end{equation}
where $\sigma$ is the sigmoid function.
$\Theta_{1,1}$ is a $1\times1$ convolution, and $\Theta_{3,3}^{2}$ is a $3\times3$ convolution with dilation rate=2.
To build the feature pyramid, we graft an up-sampling layer after the resulted features.
With the above DFAM, our framework can improve the contextual perceptions of marine animals.
\subsection{Criss-Cross Connectivity Prediction}
\begin{figure}[]
    \centering
        \resizebox{0.5\textwidth}{!}
	{
    \includegraphics[width=0.5\textwidth]{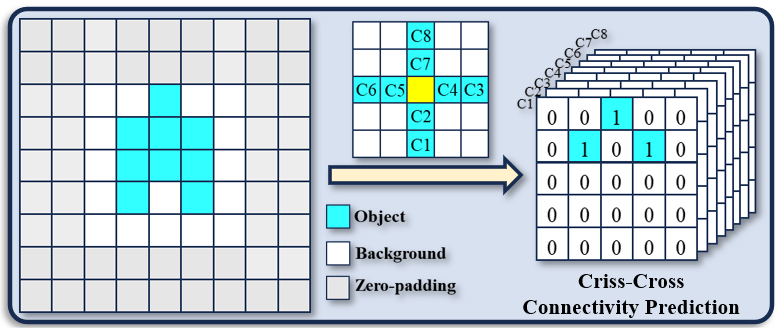}
    }
    \caption{Our Criss-Cross Connectivity Prediction (C$^3$P).}
    \vspace{-2mm}
    \label{fig:conn}
\end{figure}

Traditional image segmentation methods predict the class for each pixel.
As a result, they overlook the connectivity between discrete pixels, showing irregular structures and boundaries of objects.
To address this issue, we propose a Criss-Cross Connectivity Prediction (C$^3$P) paradigm to capture the inter-connectivity between discrete pixels.
Our approach draws inspiration from \cite{kampffmeyer2018connnet}, which emphasizes connectivity predictions between adjacent pixels.
In contrast, we extend the sampling to a criss-cross range, considering various shapes and sizes of marine animals.
Specifically, our method first transforms the single-channel mask label into an 8-channel label.
Fig.~\ref{fig:conn} illustrates these eight channels.
They represent the connectivity between their positions and the central pixel.
Given a central pixel $(w,h)$, we identify criss-cross pixels based on the following criteria:
\vspace{-2mm}
\begin{equation}
\Omega_{w, h}^{1}=\{(u, v) \||u-w|+|v-h|=1\},
\end{equation}
\vspace{-2mm}
\begin{equation}
\begin{aligned}
\Omega_{w, h}^{2}=\{(u, v) \||u-w|+|v-h|=2\\
      \cap \operatorname{Max}(|u-w|,|v-h|)=2\},
  \end{aligned}
\end{equation}
where $\Omega_{w, h}^{1}$ and $\Omega_{w, h}^{2}$are neighboring pixel sets with distances of 1 and 2, respectively.
Based on above definitions, our framework directly predict connectivity maps, which provide a more comprehensive and structured representation of segmentation masks.
The training loss function is:
\begin{equation}
    \begin{aligned}
        \mathcal{L}_{l}^{\alpha/\beta}=-\sum_{w=1}^{W} \sum_{h=1}^{H} \sum_{c=1}^{C}[Y_{l}(w, h, c) \ln(P_{l}^{\alpha/\beta}(w, h, c))\\
        +(1-Y_{l}(w, h, c))\ln (1-P_{l}^{\alpha/\beta}(w, h, c))].
    \end{aligned}
\end{equation}
Here, $P_{l}^{\alpha/\beta}$ are predicted connectivity maps at the $l$-th level from two decoders.
It is processed after the sigmoid function.
$Y_{l}$ is the corresponding ground-truth.
$(w,h)$ is the spatial location of a predicted pixel.
$c$ is the channel number.
As can be observed, our proposed C$^3$P takes the criss-cross nature of pixels and achieves vectored predictions for the animal segmentation masks.
\subsection{Pseudo-label Mutual Supervision}
To further ensure the comprehensive complementarity of dual branches, we employ the Pseudo-label Mutual Supervision (PMS) for the two decoders.
It works like a mutual learning and enables the model to optimize its parameters from a different perspective.
%
%
Specifically, we first threshold the predicted output of each level within each decoder branch.
It can be represented as follows:
\vspace{-2mm}
\begin{equation}
\hat{P}_{l}^{\alpha/\beta}=
\begin{cases}
\begin{aligned}
1, &P_{l}^{\alpha/\beta}(w, h,c)>\xi, \\
0, &otherwise.
\end{aligned}
\end{cases}
\end{equation}
where $\hat{P}_{l}^{\alpha/\beta}$ are the pseudo-labels at the $l$-th level after thresholding.
$\xi$ is the used threshold for pseudo-labels.
The above pseudo-labels are then employed to supervise the prediction of the other branch.
To this end, we use the following binary cross-entropy loss functions for training:
\vspace{-2mm}
\begin{equation}
    \begin{aligned}
        \ddot{\mathcal{L}}^{\alpha}_{l}=-\sum_{w=1}^{W} \sum_{h=1}^{H} \sum_{c=1}^{C}[\hat{P}^{\alpha}_{l}(w, h, c) \ln (\hat{P}^{\beta}_{l}(w, h, c))\\
        +(1-\hat{P}^{\alpha}_{l}(w, h, c)) \ln (1-\hat{P}^{\beta}_{l}(w, h, c))],
    \end{aligned}
\end{equation}
\vspace{-2mm}
\begin{equation}
    \begin{aligned}
        \ddot{\mathcal{L}}^{\beta}_{l}=-\sum_{w=1}^{W} \sum_{h=1}^{H} \sum_{c=1}^{C}[\hat{P}^{\beta}_{l}(w, h, c) \ln (\hat{P}^{\alpha}_{l}(w, h, c))\\
        +(1-\hat{P}^{\beta}_{l}(w, h, c)) \ln (1-\hat{P}^{\alpha}_{l}(w, h, c))].
    \end{aligned}
\end{equation}
Through the mutual supervision, we can foster a synergistic enhancement between the two branches, optimizing the extraction and integration of prompted features.

During the early stages of training, the connectivity predictions are very coarse and suboptimal.
Thus, we introduce a dynamic update coefficient for the pseudo-label supervision.
It starts at a small value, then gradually increases in an exponential manner:
\begin{equation}
\mu=0.1\times e^{-5\times\left(1-\ \frac{t}{T}\right)^2},
\end{equation}
where $t$ is the current epoch number during training.
$T$ is the total epochs.
Finally, the overall loss is expressed as:
\begin{equation}
    \mathcal{L}=\sum_{l=1}^{4}((\mathcal{L}_{l}^{\alpha}+\mathcal{L}_{l}^{\beta})+\mu(\ddot{\mathcal{L}}_{l}^{\alpha}+\ddot{\mathcal{L}}_{l}^{\beta})).
\end{equation}

For inference, we convert the connectivity maps into the binary masks.
To ensure a valid and reliable prediction, we adopt the following mutual confirmation:
\begin{equation}
P_{w, h, c}=1 \cap P_{u, v, 9-c}=1 \rightarrow P_{w,h}=1 \cap P_{u, v}=1.
\end{equation}
Thus, $P$ is the final prediction for MAS. 
\begin{table*}[]
\centering
\caption{Performance comparison on MAS3K and RMAS. The best and second results are in red and blue, respectively.}
\label{mask3k_res}
\vspace{-2mm}
\renewcommand\arraystretch{1.1}
\setlength\tabcolsep{5.5pt}
\resizebox{0.80\textwidth}{!}
{
\begin{tabular}{c|c|c|c|c|c|c|c|c|c|c}
\hline
&\multicolumn{5}{c|}{\textbf{MAS3K}}        &\multicolumn{5}{c}{\textbf{RMAS}} \\ \cline{2-11}
\textbf{Method}&\textbf{mIoU} & $\textbf{S}_\alpha$ & $\textbf{F}_\beta^w $&$ \textbf{mE}_\phi$ & \textbf{MAE}& \textbf{mIoU} & $\textbf{S}_\alpha$ & $\textbf{F}_\beta^w$ & $\textbf{mE}_\phi$ & \textbf{MAE} \\
\hline
 SINet~\cite{fan2020camouflaged} &0.658&0.820&0.725&0.884&0.039&0.684&0.835&0.780&0.908&0.025\\
 PFNet~\cite{mei2021camouflaged}  &0.695&0.839&0.746&0.890&0.039&0.694&0.843&0.771&0.922&0.026\\
 RankNet~\cite{lv2021simultaneously} &0.658&0.812&0.722&0.867&0.043&0.704&0.846&0.772&0.927&0.026\\
 C2FNet~\cite{sun2021context}   &0.717&0.851&0.761&0.894&0.038&0.721&0.858&0.788&0.923&0.026\\
 ECDNet~\cite{li2021marine}   &0.711&0.850&0.766&0.901&0.036&0.664&0.823&0.689&0.854&0.036\\
 OCENet~\cite{liu2022modeling}  &0.667&0.824&0.703&0.868&0.052&0.680&0.836&0.752&0.900&0.030\\
 ZoomNet~\cite{pang2022zoom}   &0.736&0.862&0.780&0.898&0.032&0.728&0.855&0.795&0.915&0.022\\
 MASNet~\cite{fu2023masnet} &0.742&0.864&0.788&0.906&0.032&\textcolor{blue}{\textbf{0.731}}&\textcolor{red}{\textbf{0.862}}&\textcolor{blue}{\textbf{0.801}}&0.920&0.024\\
    \hline
    SETR~\cite{zheng2021rethinking}   &0.715&0.855&0.789&0.917&0.030&0.654&0.818&0.747&0.933&0.028\\
    TransUNet~\cite{chen2021transunet}   &0.739&0.861&0.805&0.919&0.029&0.688&0.832&0.776&\textcolor{blue}{\textbf{0.941}}&0.025\\

    H2Former~\cite{he2023h2former}   &\textcolor{blue}{\textbf{0.748}}&0.865&\textcolor{blue}{\textbf{0.810}}&\textcolor{blue}{\textbf{0.925}}&\textcolor{blue}{\textbf{0.028}}&0.717&0.844&0.799&0.931&\textcolor{blue}{\textbf{0.023}}\\
    \hline
    SAM~\cite{kirillov2023segment}  &0.566&0.763&0.656&0.807&0.059&0.445&0.697&0.534&0.790&0.053\\
    SAM-Ad\cite{chen2023sam}   &0.714&0.847&0.782&0.914&0.033&0.656&0.816&0.752&0.927&0.027\\
    SAM-DA~\cite{lai2023detect}   &0.742&\textcolor{blue}{\textbf{0.866}}&0.806&0.925&0.028&0.686&0.833&0.780&0.926&0.024\\
    \hline
\textbf{Dual-SAM}   &\textcolor{red}{\textbf{0.789}}&\textcolor{red}{\textbf{0.884}}&\textcolor{red}{\textbf{0.838}}&\textcolor{red}{\textbf{0.933}}&\textcolor{red}{\textbf{0.023}}&\textcolor{red}{\textbf{0.735}}&\textcolor{blue}{\textbf{0.860}}&\textcolor{red}{\textbf{0.812}}&\textcolor{red}{\textbf{0.944}}&\textcolor{red}{\textbf{0.022}}\\
\hline
\end{tabular}
}
\end{table*}
\begin{table*}[h]
\centering
\caption{Performance comparison on UFO120 and RUWI. The best and second results are in red and blue, respectively.}
\label{ufo_res}
\vspace{-2mm}
\renewcommand\arraystretch{1.1}
\setlength\tabcolsep{5.5pt}
    \resizebox{0.80\textwidth}{!}
	{
\begin{tabular}{c|c|c|c|c|c|c|c|c|c|c}
        \hline
        &\multicolumn{5}{c|}{\textbf{UFO120}}        &\multicolumn{5}{c}{\textbf{RUWI}} \\ \cline{2-11}
    \textbf{Method}&\textbf{mIoU} & $\textbf{S}_\alpha$ & $\textbf{F}_\beta^w$ & $\textbf{m}\textbf{E}_\phi$ & \textbf{MAE}& \textbf{mIoU} & $\textbf{S}_\alpha$ & $\textbf{F}_\beta^w$ & $\textbf{m}\textbf{E}_\phi$ & \textbf{MAE} \\
        \hline

    SINet~\cite{fan2020camouflaged}   &0.767&0.837&0.834&0.890&0.079&0.785&0.789&0.825&0.872&0.096\\
    PFNet~\cite{mei2021camouflaged}   &0.570&0.708&0.550&0.683&0.216&0.864&0.883&0.870&0.790&0.062\\
    RankNet~\cite{lv2021simultaneously}   &0.739&0.823&0.772&0.828&0.101&0.865&0.886&0.889&0.759&0.056\\
    C2FNet~\cite{sun2021context}   &0.747&0.826&0.806&0.878&0.083&0.840&0.830&0.883&0.924&0.060\\
    ECDNet~\cite{li2021marine}   &0.693&0.783&0.768&0.848&0.103&0.829&0.812&0.871&0.917&0.064\\
    OCENet~\cite{liu2022modeling}  &0.605&0.725&0.668&0.773&0.161&0.763&0.791&0.798&0.863&0.115\\
    ZoomNet~\cite{pang2022zoom}   &0.616&0.702&0.670&0.815&0.174&0.739&0.753&0.771&0.817&0.137\\
    MASNet~\cite{fu2023masnet} &0.754&0.827&0.820&0.879&0.083&0.865&0.880&0.913&0.944&0.047\\
    \hline

    SETR~\cite{zheng2021rethinking}   &0.711&0.811&0.796&0.871&0.089&0.832&0.864&0.895&0.924&0.055\\
    TransUNet~\cite{chen2021transunet}   &0.752&0.825&0.827&0.888&0.079&0.854&0.872&0.910&0.940&0.048\\

    H2Former~\cite{he2023h2former}   &\textcolor{blue}{\textbf{0.780}}&\textcolor{blue}{\textbf{0.844}}&\textcolor{blue}{\textbf{0.845}}&\textcolor{blue}{\textbf{0.901}}&\textcolor{blue}{\textbf{0.070}}&0.871&0.884&0.919&0.945&0.045\\
    \hline
    SAM~\cite{kirillov2023segment}   &0.681&0.768&0.745&0.827&0.121&0.849&0.855&0.907&0.929&0.057\\
    SAM-Ad~\cite{chen2023sam}   &0.757&0.829&0.834&0.884&0.081&0.867&0.878&0.913&\textcolor{blue}{\textbf{0.946}}&0.046\\
    SAM-DA~\cite{lai2023detect}   &0.768&0.841&0.836&0.893&0.073&\textcolor{blue}{\textbf{0.881}}&\textcolor{blue}{\textbf{0.889}}&\textcolor{blue}{\textbf{0.925}}&0.940&\textcolor{blue}{\textbf{0.044}}\\
    \hline
\textbf{Dual-SAM}   &\textcolor{red}{\textbf{0.810}}&\textcolor{red}{\textbf{0.856}}&\textcolor{red}{\textbf{0.864}}&\textcolor{red}{\textbf{0.914}}&\textcolor{red}{\textbf{0.064}}&\textcolor{red}{\textbf{0.904}}&\textcolor{red}{\textbf{0.903}}&\textcolor{red}{\textbf{0.939}}&\textcolor{red}{\textbf{0.959}}&\textcolor{red}{\textbf{0.035}}\\
\hline
\end{tabular}
}
\vspace{-4mm}
\end{table*}
\section{Experiments}
\subsection{Datasets and Evaluation Metrics}
To thoroughly validate the performance, we adopt five public datasets and five evaluation metrics.

For the datasets, \textbf{MAS3K}~\cite{li2020mas3k} contains 3,103 images with high-quality annotations.
We follow the default split and use 1,769 images for training and 1,141 images for testing.
We exclude 193 images that only have a background.
\textbf{RMAS}~\cite{fu2023masnet} includes 3,014 marine images.
We use 2,514 images for training and 500 images for testing.
\textbf{UFO120}~\cite{islam2020simultaneous} contains a total of 1,620 marine images.
We use 1,500 images for training and 120 images for testing.
\textbf{RUWI}~\cite{drews2021underwater} contains 700 marine images.
We use 525 images for training and 175 images for testing.
In addition, to verify the generalization, we adopt the \textbf{USOD10K}~\cite{hong2023usod10k} dataset.
It is the largest underwater salient object detection dataset with a total of 10,255 images, splitting 9,229 images for training and 1,026 images for testing.

To evaluate the model's performance, we utilize the following five metrics: Mean Intersection over Union ($mIoU$), Structural Similarity Measure ($S_\alpha$), Weighted F-measure ($F_\beta^w$), Mean Enhanced-Alignment Measure ($m\textbf{E}_\phi$), Mean Absolute Error ($MAE$).
These metrics offer a comprehensive evaluation, capturing various aspects of segmentation quality.
For more details on these metrics, please refer to the supplementary material.
\subsection{Implementation Details}
We implement our model with the PyTorch toolbox and conduct experiments with one RTX 3090 GPU.
In our model, the SAM's encoder is initialized from the pre-trained SAM-B~\cite{kirillov2023segment}, while the rest are randomly initialized.
During the training process, we freeze the SAM's encoder and only fine-tune the remaining modules.
To reduce the computation, we set $j=3\times i$ for the MCP.
The threshold $\xi$ is set to 0.5.
The AdamW optimizer~\cite{loshchilov2017decoupled} is used to update the parameters.
The initial learning rate and weight decay are set to 0.001 and 0.1, respectively.
We reduce the learning rate by a factor of 10 at every 20 epochs.
The total number of training epochs $T$ is set to 50.
The mini-batch size is set to 8 due to the memory limitation.
All the input images are resized to $512\times512\times3$.
For the evaluation, we resize the predicted masks back to the original image size by the bilinear interpolation.
\begin{figure*}[!t]
\centering
\resizebox{0.88\textwidth}{!}
{
\renewcommand\arraystretch{0.1}
\begin{tabular}{@{}c@{}c@{}c@{}c@{}c@{}c@{}c@{}c@{}c@{}c@{}c@{}c@{}c@{}c@{}c@{}c}

\vspace{0.5mm}
\includegraphics[width=0.0909\linewidth,height=1.6cm]{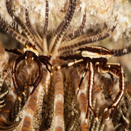}\ &
\includegraphics[width=0.0909\linewidth,height=1.6cm]{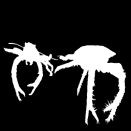}\ &
\includegraphics[width=0.0909\linewidth,height=1.6cm]{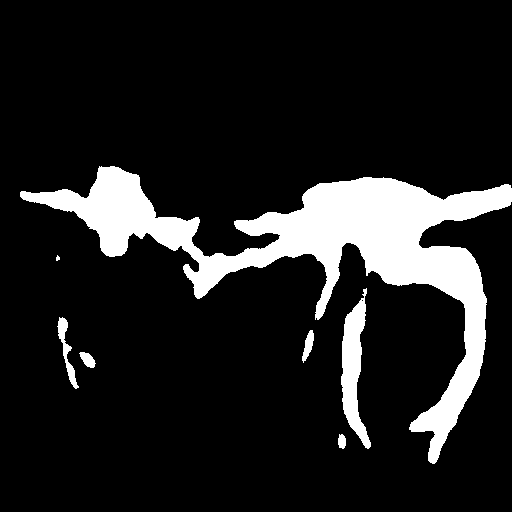}\ &
\includegraphics[width=0.0909\linewidth,height=1.6cm]{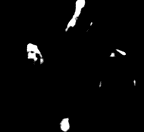}\ &
\includegraphics[width=0.0909\linewidth,height=1.6cm]{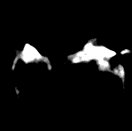}\ &
\includegraphics[width=0.0909\linewidth,height=1.6cm]{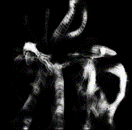}\ &
\includegraphics[width=0.0909\linewidth,height=1.6cm]{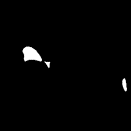}\ &
\includegraphics[width=0.0909\linewidth,height=1.6cm]{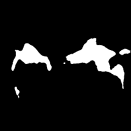}\ &
\includegraphics[width=0.0909\linewidth,height=1.6cm]{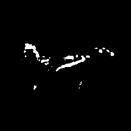}\ &
\includegraphics[width=0.0909\linewidth,height=1.6cm]{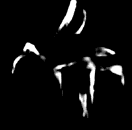}\ &
\includegraphics[width=0.0909\linewidth,height=1.6cm]{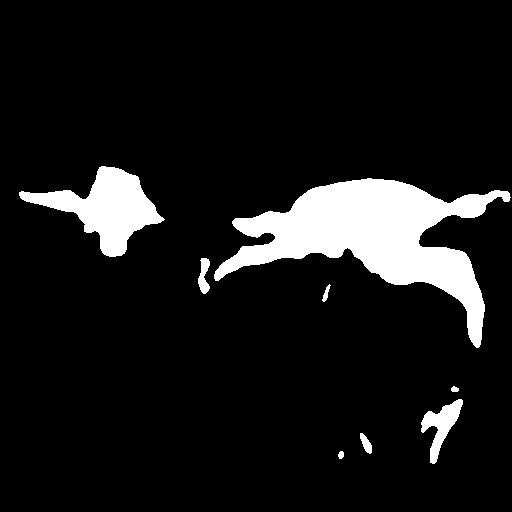}\  \\

\vspace{0.5mm}
\includegraphics[width=0.0909\linewidth,height=1.6cm]{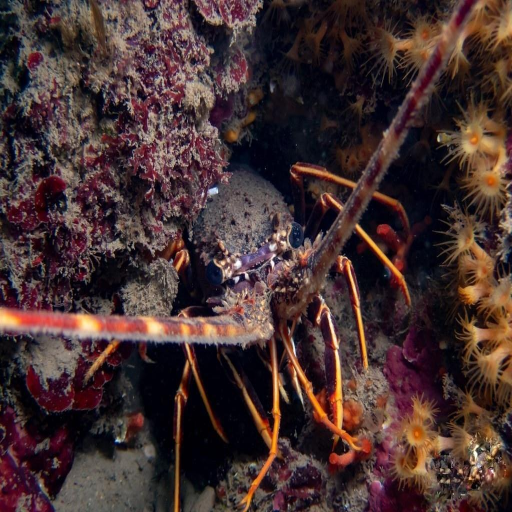}\ &
\includegraphics[width=0.0909\linewidth,height=1.6cm]{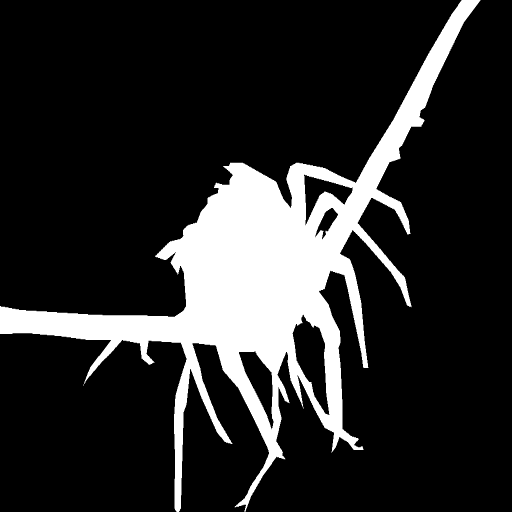}\ &
\includegraphics[width=0.0909\linewidth,height=1.6cm]{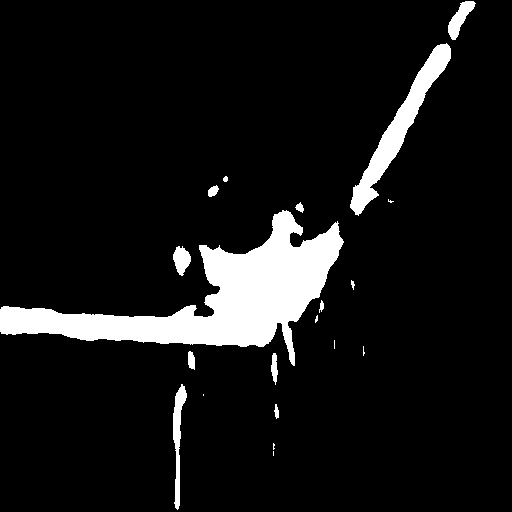}\ &
\includegraphics[width=0.0909\linewidth,height=1.6cm]{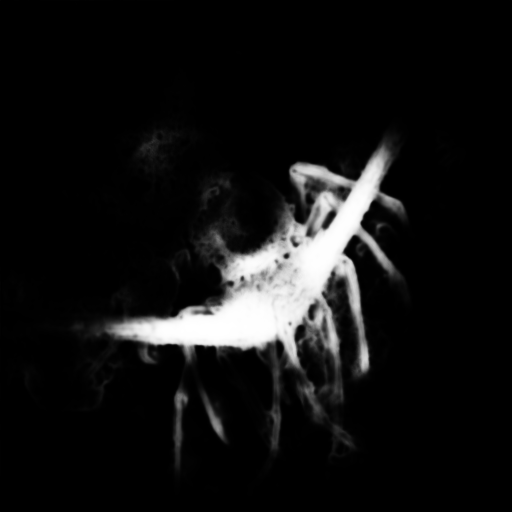}\ &
\includegraphics[width=0.0909\linewidth,height=1.6cm]{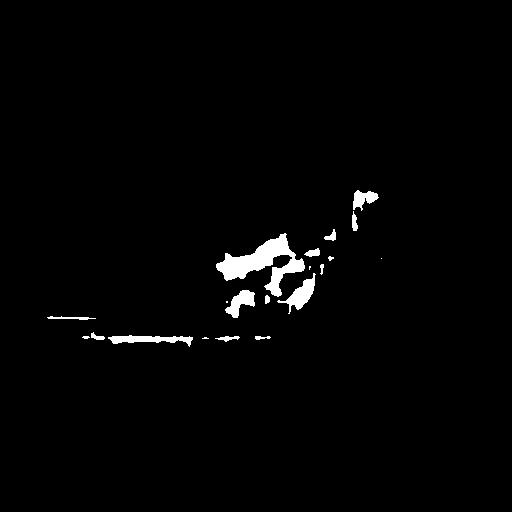}\ &
\includegraphics[width=0.0909\linewidth,height=1.6cm]{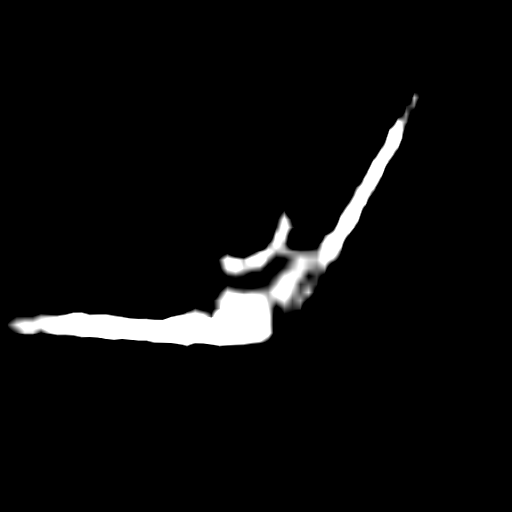}\ &
\includegraphics[width=0.0909\linewidth,height=1.6cm]{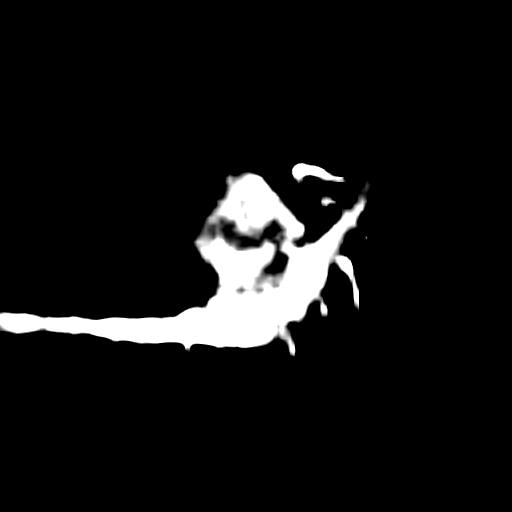}\ &
\includegraphics[width=0.0909\linewidth,height=1.6cm]{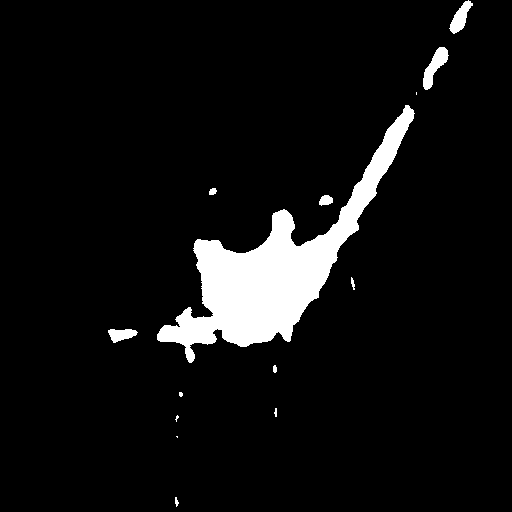}\ &
\includegraphics[width=0.0909\linewidth,height=1.6cm]{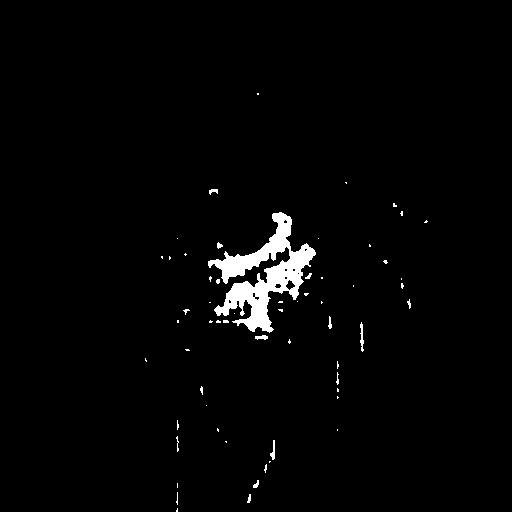}\ &
\includegraphics[width=0.0909\linewidth,height=1.6cm]{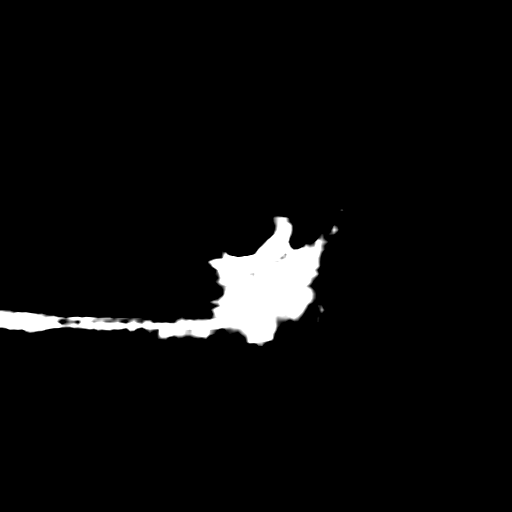}\ &
\includegraphics[width=0.0909\linewidth,height=1.6cm]{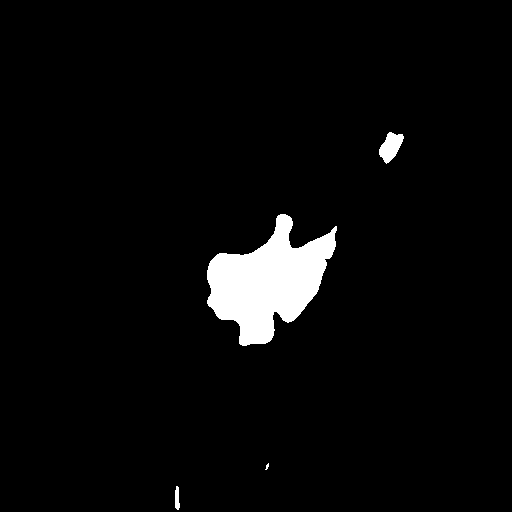}\  \\
\vspace{0.5mm}
\includegraphics[width=0.0909\linewidth,height=1.6cm]{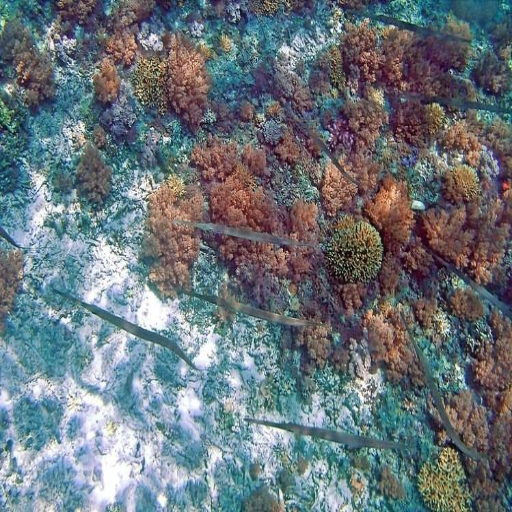}\ &
\includegraphics[width=0.0909\linewidth,height=1.6cm]{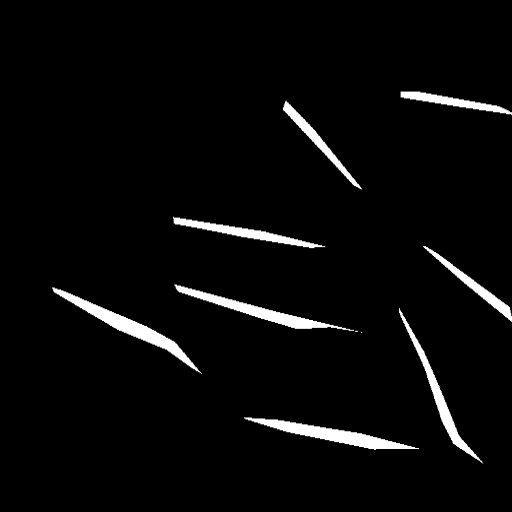}\ &
\includegraphics[width=0.0909\linewidth,height=1.6cm]{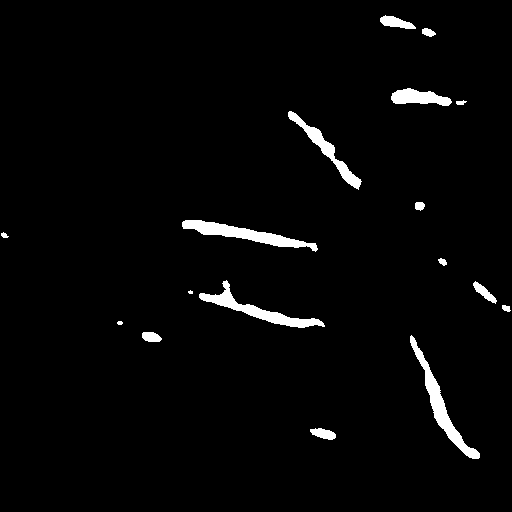}\ &
\includegraphics[width=0.0909\linewidth,height=1.6cm]{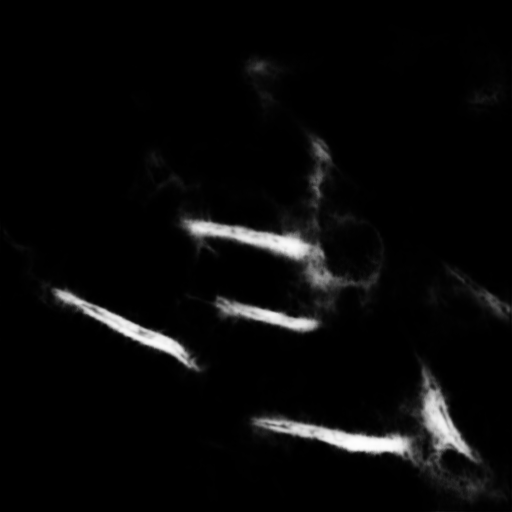}\ &
\includegraphics[width=0.0909\linewidth,height=1.6cm]{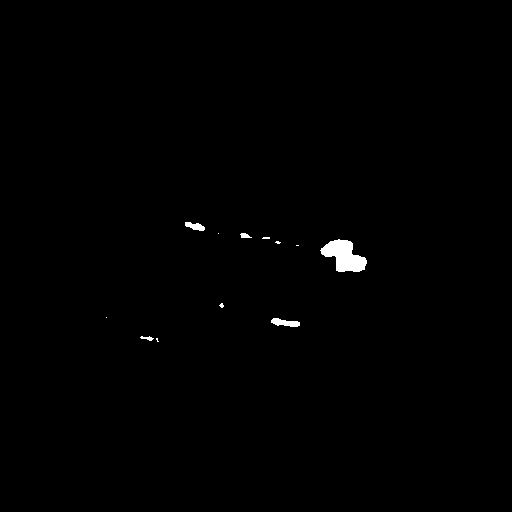}\ &
\includegraphics[width=0.0909\linewidth,height=1.6cm]{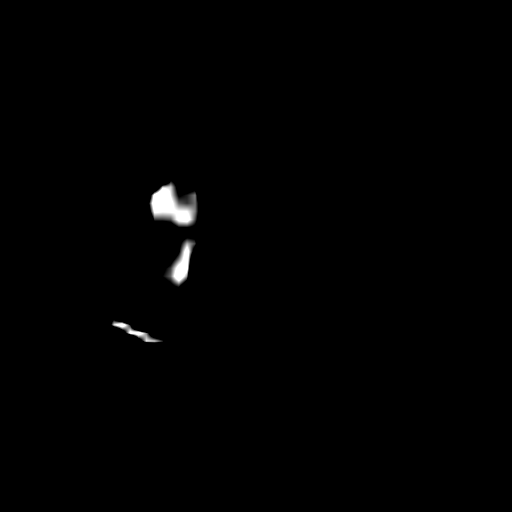}\ &
\includegraphics[width=0.0909\linewidth,height=1.6cm]{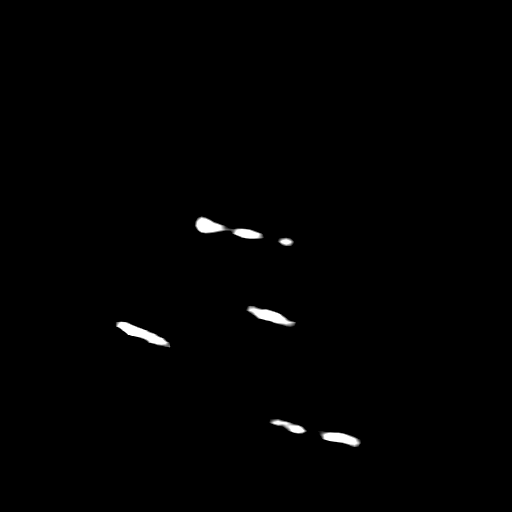}\ &
\includegraphics[width=0.0909\linewidth,height=1.6cm]{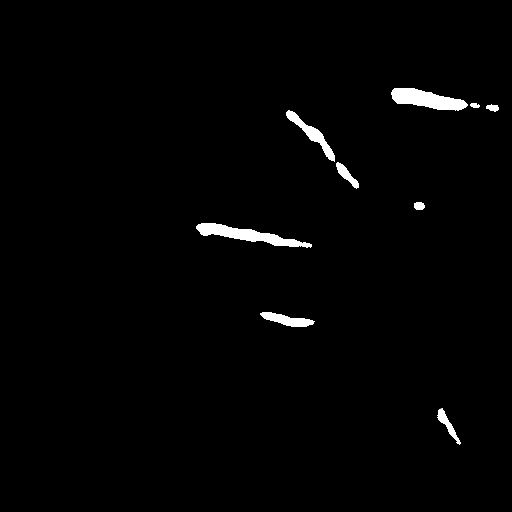}\ &
\includegraphics[width=0.0909\linewidth,height=1.6cm]{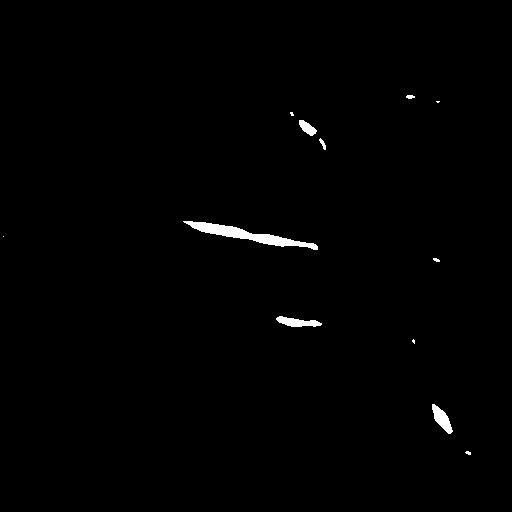}\ &
\includegraphics[width=0.0909\linewidth,height=1.6cm]{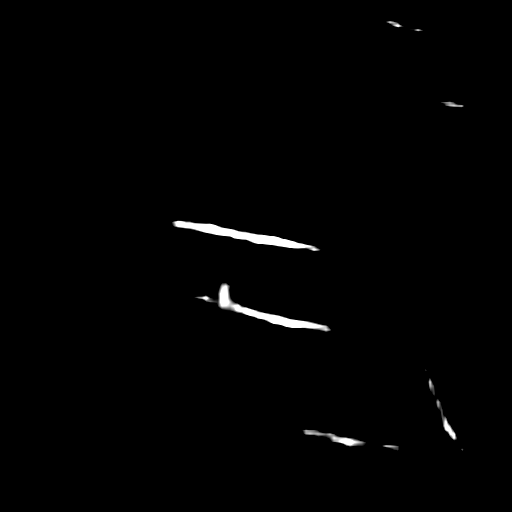}\ &
\includegraphics[width=0.0909\linewidth,height=1.6cm]{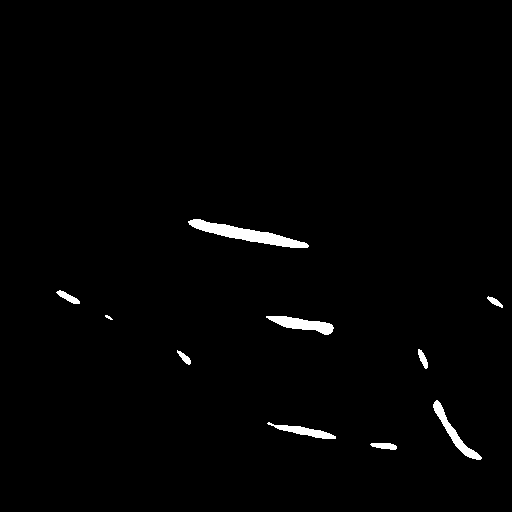}\  \\

\vspace{0.5mm}
\includegraphics[width=0.0909\linewidth,height=1.6cm]{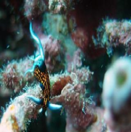}\ &
\includegraphics[width=0.0909\linewidth,height=1.6cm]{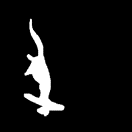}\ &
\includegraphics[width=0.0909\linewidth,height=1.6cm]{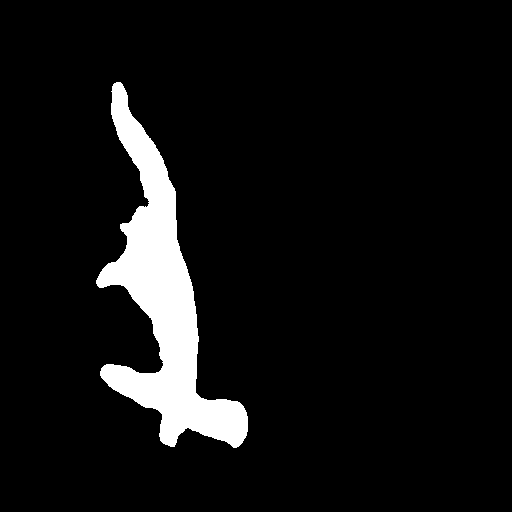}\ &
\includegraphics[width=0.0909\linewidth,height=1.6cm]{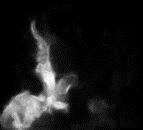}\ &
\includegraphics[width=0.0909\linewidth,height=1.6cm]{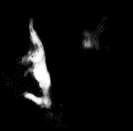}\ &
\includegraphics[width=0.0909\linewidth,height=1.6cm]{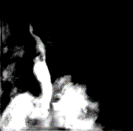}\ &
\includegraphics[width=0.0909\linewidth,height=1.6cm]{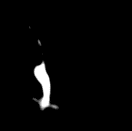}\ &
\includegraphics[width=0.0909\linewidth,height=1.6cm]{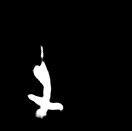}\ &
\includegraphics[width=0.0909\linewidth,height=1.6cm]{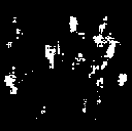}\ &
\includegraphics[width=0.0909\linewidth,height=1.6cm]{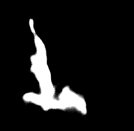}\ &
\includegraphics[width=0.0909\linewidth,height=1.6cm]{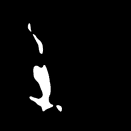}\  \\

\vspace{0.5mm}
\includegraphics[width=0.0909\linewidth,height=1.6cm]{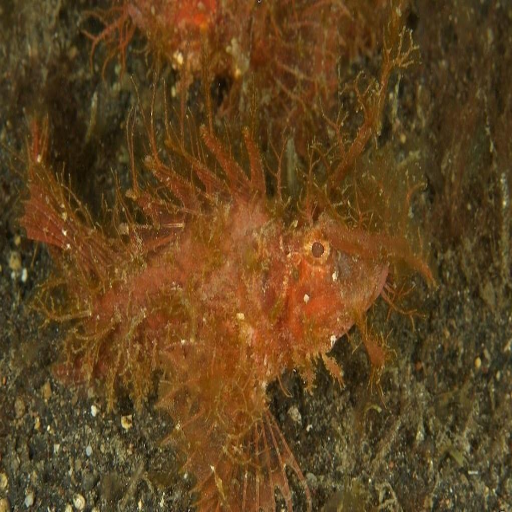}\ &
\includegraphics[width=0.0909\linewidth,height=1.6cm]{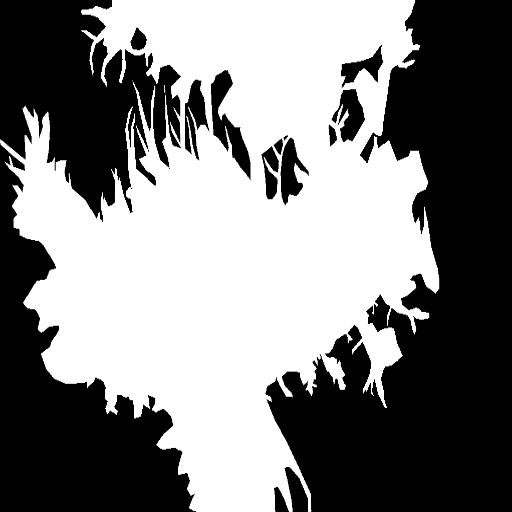}\ &
\includegraphics[width=0.0909\linewidth,height=1.6cm]{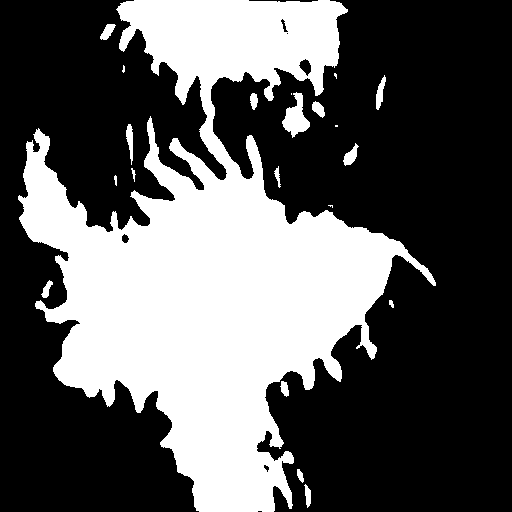}\ &
\includegraphics[width=0.0909\linewidth,height=1.6cm]{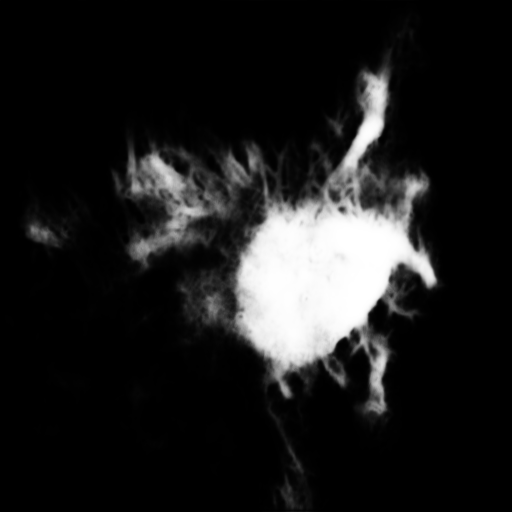}\ &
\includegraphics[width=0.0909\linewidth,height=1.6cm]{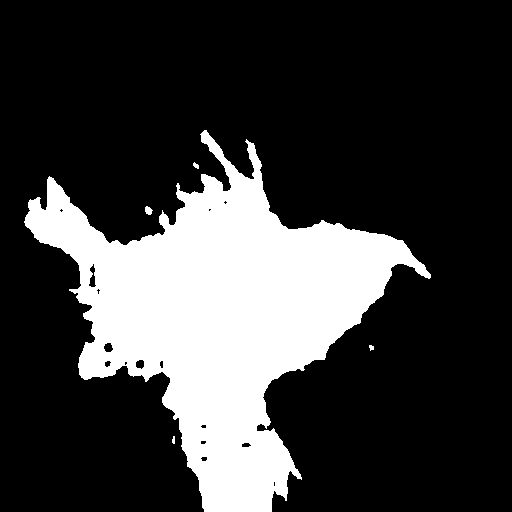}\ &
\includegraphics[width=0.0909\linewidth,height=1.6cm]{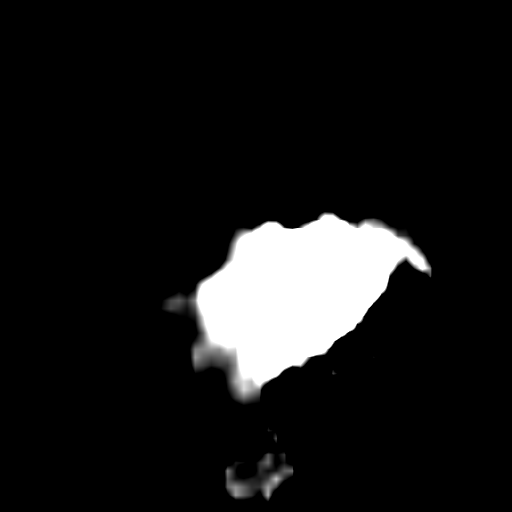}\ &
\includegraphics[width=0.0909\linewidth,height=1.6cm]{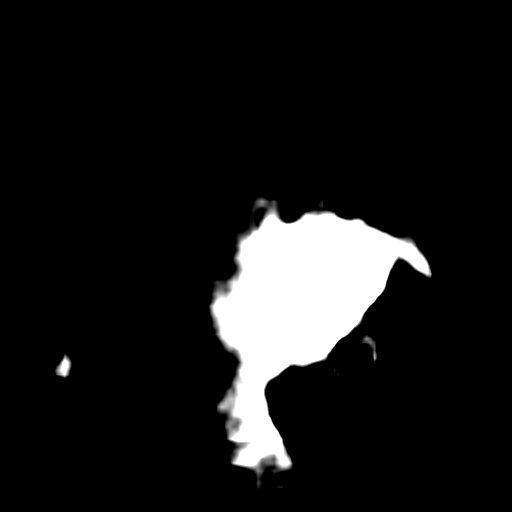}\ &
\includegraphics[width=0.0909\linewidth,height=1.6cm]{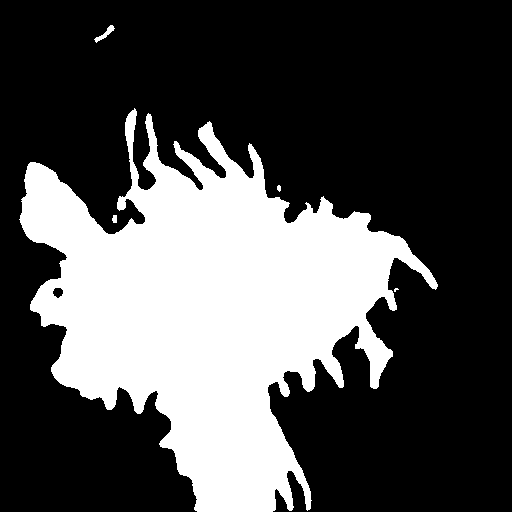}\ &
\includegraphics[width=0.0909\linewidth,height=1.6cm]{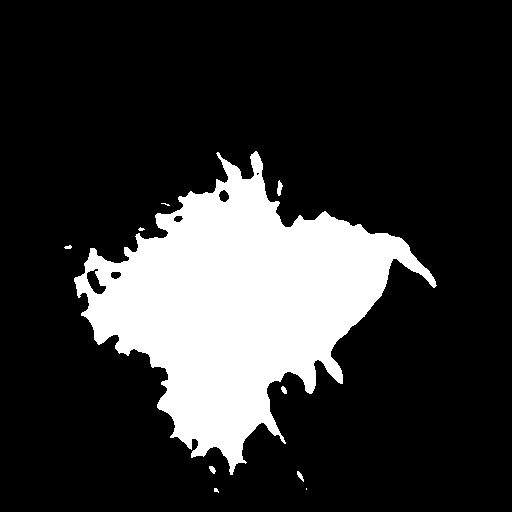}\ &
\includegraphics[width=0.0909\linewidth,height=1.6cm]{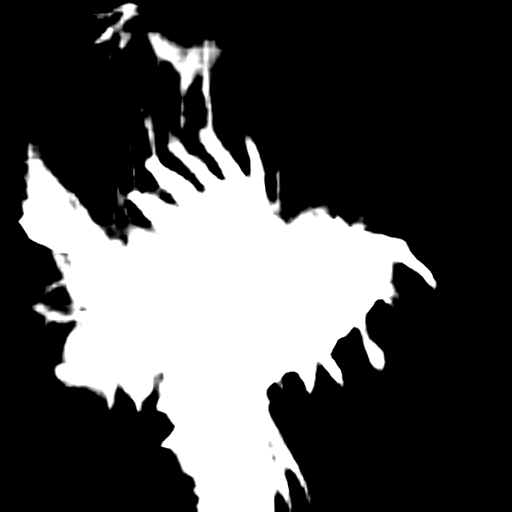}\ &
\includegraphics[width=0.0909\linewidth,height=1.6cm]{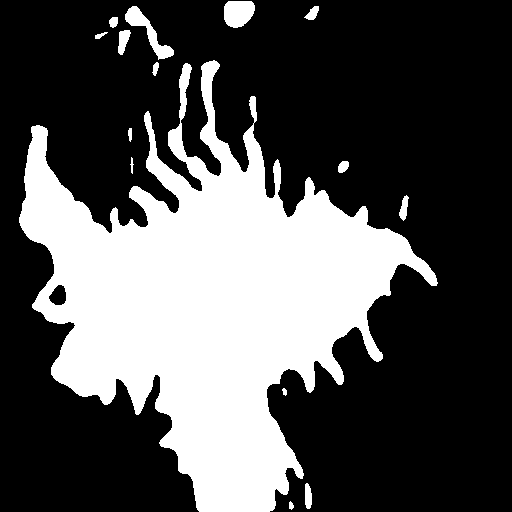}\  \\

\vspace{0.5mm}
\includegraphics[width=0.0909\linewidth,height=1.6cm]{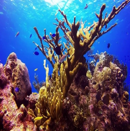}\ &
\includegraphics[width=0.0909\linewidth,height=1.6cm]{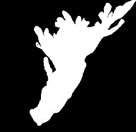}\ &
\includegraphics[width=0.0909\linewidth,height=1.6cm]{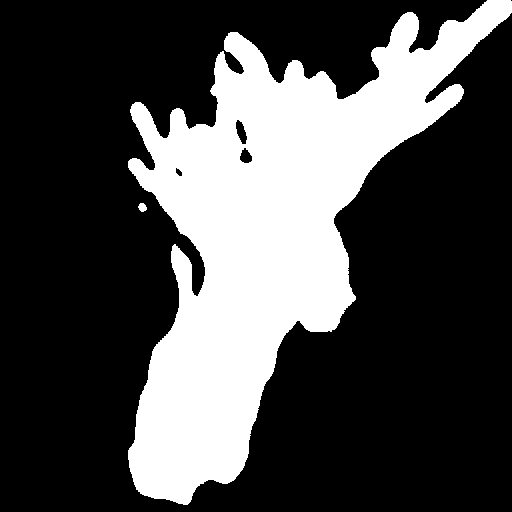}\ &
\includegraphics[width=0.0909\linewidth,height=1.6cm]{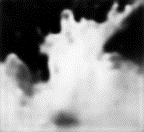}\ &
\includegraphics[width=0.0909\linewidth,height=1.6cm]{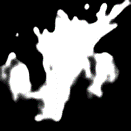}\ &
\includegraphics[width=0.0909\linewidth,height=1.6cm]{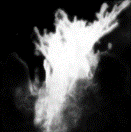}\ &
\includegraphics[width=0.0909\linewidth,height=1.6cm]{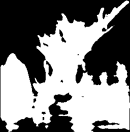}\ &
\includegraphics[width=0.0909\linewidth,height=1.6cm]{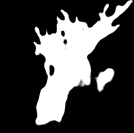}\ &
\includegraphics[width=0.0909\linewidth,height=1.6cm]{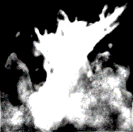}\ &
\includegraphics[width=0.0909\linewidth,height=1.6cm]{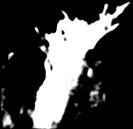}\ &
\includegraphics[width=0.0909\linewidth,height=1.6cm]{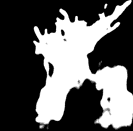}\  \\
 {\small Image} & {\small GT} & {\small Dual-SAM} & {\small ECDNet}& {\small MASNet}  & {\small SETR} & {\small TransUNet} & {\small H2Former} & {\small SAM} & {\small SAM-Ad}& {\small SAM-DA} \\
\end{tabular}
}
\vspace{-2mm}
\caption{Visual comparison of predicted segmentation masks with different methods.}
\vspace{-2mm}
\label{fig:visual}
\end{figure*}
\begin{table}[]
\centering
\caption{Performance comparison on USOD10k. The best and second results are in red and blue, respectively.}
\vspace{-2mm}
\resizebox{0.46\textwidth}{!}
{
\begin{tabular}{c|c|c|c|c}
\hline
&\multicolumn{4}{c}{\textbf{USOD10K}}        \\ \cline{2-5}
\textbf{Method}& $\textbf{S}_\alpha$ & $\textbf{m}\textbf{E}_\phi$ & $\textbf{maxF}$ & \textbf{MAE}\\ \hline
S2MA ~\cite{liu2021learning}                    & .8664                     & .9208 & .8530 & .0558 \\
SGL-KRN ~\cite{xu2021locate}                 & .9214     & .9633 & \textcolor{blue}{\textbf{.9245}} & .0237 \\
DCF ~\cite{ji2021calibrated}                     & .9116                     & .9541 & .9045 & .0312 \\
SPNet ~\cite{zhou2021specificity}                   & .9075                     & .9554 & .9069 & .0280 \\
HAINet  ~\cite{li2021hierarchical}                 & .9123                     & .9552 & .9116 & .0279 \\
VST ~\cite{liu2021swin}                      & .9136                     & .9614 & .9108 & .0267\\
TriTransNet ~\cite{liu2021tritransnet}              & .7889                     & .8479 & .7501 & .0659 \\
CSNet ~\cite{cheng2021highly}                   & .8595                     & .9178 & .8462 & .0548 \\
D3Net ~\cite{fan2020rethinking}                   & .8931                     & .9413 & .8807 & .0374 \\
SVAM-Net ~\cite{islam2020svam}                & .7465                     &.7649  & .6451 & .0915\\
BTS-Net ~\cite{zhang2021bts}                  & .9093                     & .9542 & .9104 & .0291 \\
CDINet ~\cite{zhang2021cross}                  & .7049                     & .8644 & .7362 & .0904 \\
CTDNet  ~\cite{zhao2021complementary}                 & .9085                     & .9531 & .9073 & .0285   \\
MFNet ~\cite{piao2021mfnet}                   & .8425                     & .9146 & .8193 & .0512   \\
PFSNet ~\cite{ma2021pyramidal}                  & .8983                     & .9421 & .8966 & .0370 \\
PSGLoss ~\cite{yang2021progressive}                 & .8640                     & .9078 & .8508 & .0417 \\
TC-USOD~\cite{hong2023usod10k}                  & \textcolor{blue}{\textbf{.9215}}&\textcolor{blue}{\textbf{.9683}}&.9236&\textcolor{blue}{\textbf{.0201}}\\
\hline
SAM~\cite{kirillov2023segment}&.8543&.9095&.8812&.0380\\
SAM-Ad~\cite{chen2023sam}&.8952&.9533&.9153&.0276\\
SAM-DA~\cite{lai2023detect}&.9051&.9552&.9154&.0250\\\hline
\textbf{Dual-SAM}       &\textcolor{red}{\textbf{.9238} }   & \textcolor{red}{\textbf{.9684}} &\textcolor{red}{\textbf{.9311}}  &\textcolor{red}{\textbf{.0185}}\\ \hline
\end{tabular}
}
\vspace{-4mm}
\label{USOD10k}
\end{table}
\subsection{Comparisons with the State-of-the-arts}
In this part, we compare our method with other methods.
The quantitative and qualitative results clearly show the notable advantage of our proposed method.

\textbf{Quantitative Comparisons.}
Tab.~\ref{mask3k_res} and Tab.~\ref{ufo_res} show quantitative comparisons on typical MAS datasets.
%
%
%
When compared with CNN-based methods, our method notably improves the performance.
On the challenging MAS3K dataset, our method achieves the highest scores across all metrics.
It delivers a 3-5\% improvement in various metrics.
Meanwhile, our method consistently performs better on other MAS datasets.
When compared with Transformer-based methods, our method delivers a 2-3\% improvement on the MAS3K dataset.
%
%
When compared with other SAM-based methods, our method shows a 3-4\% boost in performance.
Besides, in Tab.~\ref{USOD10k}, we compare our method with other methods for underwater salient object detection.
Our proposed method still achieves excellent results.

\textbf{Qualitative Comparisons.}
Fig.~\ref{fig:visual} shows some visual examples to further verify the effectiveness of our method.
As can be observed, our method can obtain better results in terms of whole structures (the 1st-2nd rows), multiple animals (the 3rd row), camouflage animals (the 4th row) and fine-grained boundaries (the 5th-6th rows).
When compared with other SAM-based methods, our method can consistently improve the performance.
The main reason is that our method introduces effective prompts and decoders.
%
\subsection{Ablation Study}
In this subsection, we conduct experiments to analyse the effect of different modules.
The results are reported on the MAS3K dataset.
Similar trends appear on other datasets.

\textbf{Effect of Different Mask Prediction Paradigms.}
Tab.~\ref{table:abla_1} shows the segmentation performance with different mask prediction paradigms.
Clearly, the connectivity prediction delivers superior performance than the pixel-wise prediction.
in predicting both pixel-level connectivity and vector-level connectivity.
Our proposed C$^3$P consistently shows better results than the connectivity prediction method~\cite{kampffmeyer2018connnet} and pixel-wise prediction.
It indicates a more comprehensive understanding of marine animals.
\begin{table}[h!]
\centering
\caption{Performance comparison of different prediction methods. }
\label{table:abla_1}
\vspace{-2mm}
\resizebox{0.48\textwidth}{!}
{
\begin{tabular}{l|c|c|c|c|c}
\hline
Method & mIoU & $S_{\alpha}$ & $F^w_{\beta}$ & $mE_{\phi}$ & MAE \\
\hline
Pixel-wise & 0.772 & 0.875 & 0.825 & 0.923 & 0.027 \\
Nearby~\cite{kampffmeyer2018connnet} & 0.781 & 0.879 & 0.829 & 0.929 & 0.026 \\
C$^3$P (Ours)   & 0.789 & 0.884 & 0.838 & 0.933 & 0.023 \\
\hline
\end{tabular}
}
\end{table}

\textbf{Effect of Dual Branches.}
In this work, we introduce dual branches to improve the ability of SAM for MAS.
Tab.~\ref{table:abla_2} shows the performance comparison.
As can be observed, the model with dual branches achieves better results than the single branch across all the metrics.
It clearly demonstrates the effectiveness of our dual structures for marine feature extraction.

\begin{figure}
    \centering
        \resizebox{0.48\textwidth}{!}
	{
    \includegraphics[width=0.5\textwidth]{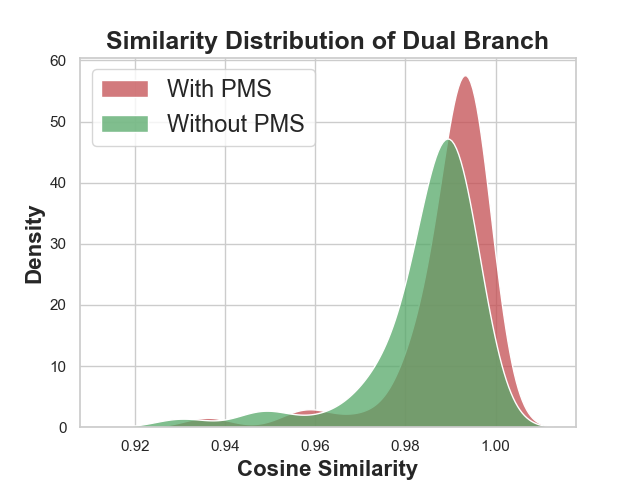}
    }
    \vspace{-4mm}
    \caption{Complementary effects of PMS on dual branch results.}
    \label{fig:f4}
    \vspace{-4mm}
\end{figure}

\textbf{Effect of PMS.}
In this work, we employ PMS to further ensure the comprehensive complementarity of dual branches.
Tab.~\ref{table:abla_2} shows the performance comparison.
In addition, Fig.~\ref{fig:f4} illustrates effects of the PMS.
As can be observed, the performance is significantly improved by incorporating the PMS.
The PMS can achieve a complementary effect in predicting segmentation masks.
\begin{table}[h!]
\centering
\caption{Performance comparison of dual branches and PMS.}
\vspace{-2mm}
    \resizebox{0.48\textwidth}{!}
	{
\begin{tabular}{l|c|c|c|c|c}
\hline
Method & mIoU & $S_{\alpha}$ & $F^w_{\beta}$ & $mE_{\phi}$ & MAE \\
\hline
Single Branch & 0.767 & 0.872 & 0.816 & 0.922 & 0.028 \\
Dual w/o PMS & 0.771 & 0.874 & 0.820 & 0.923 & 0.029 \\
Dual w PMS   & 0.789 & 0.884 & 0.838 & 0.933 & 0.023 \\
\hline
\end{tabular}
}
\label{table:abla_2}
\end{table}

\textbf{Effect of MCP.}
In this work, we inject multi-level prompt information into SAM's encoder for prior guidance.
Tab.~\ref{table:abla_3} shows the performance effect of MCP.
With the proposed MCP, the model can improve the performances across all the metrics.
The main reason is that the MCP helps SAM'encoder incorporate more fine-grained information.
\begin{table}[h!]
\centering
\caption{Performance effect of MCP.}
\vspace{-2mm}
\resizebox{0.48\textwidth}{!}
{
\begin{tabular}{l|c|c|c|c|c}
\hline
Method & mIoU & $S_{\alpha}$ & $F^w_{\beta}$ & $mE_{\phi}$ & MAE \\
\hline
w/o MCP & 0.778 & 0.877 & 0.825 & 0.929 & 0.026 \\
w MCP   & 0.789 & 0.884 & 0.838 & 0.933 & 0.023 \\
\hline
\end{tabular}
}
\label{table:abla_3}
\end{table}

\textbf{Effect of DFAM.}
In this work, we propose DFAM to fuse the prompted features.
Tab.~\ref{table:abla_4} provides the performance effect of DFAM.
With the proposed MCP, the model can improve the performances across all the metrics, especially in mIoU and MAE
In fact, the improved results mainly come from the dilated convolution and channel attention, which aggregate both semantic and detail information.
\begin{table}[h!]
\centering
\caption{Performance effect of DFAM.}
\vspace{-2mm}
    \resizebox{0.48\textwidth}{!}
	{
\begin{tabular}{l|c|c|c|c|c}
\hline
Method & mIoU & $S_{\alpha}$ & $F^w_{\beta}$ & $mE_{\phi}$ & MAE \\
\hline
w/o DFAM & 0.769 & 0.873 & 0.821 & 0.921 & 0.028 \\
w DFAM   & 0.789 & 0.884 & 0.838 & 0.933 & 0.023 \\
\hline
\end{tabular}
}
\label{table:abla_4}
\end{table}

\textbf{Effect of Adapters.}
In this work, we introduce multiple adapters into the SAM's encoder for model adaptation.
Tab.~\ref{table:abla_5} shows the effectiveness of different adapter mechanisms.
As can be observed, the performance shows a considerable decrease when removing these adapters.
These adapters play a crucial role for extracting domain-specific features.
The adapters have a significant impact on each subsequent module. From the experimental results, it is evident that both types of adapters we employ can substantially and efficiently enhance the model's performance.
\begin{table}[h!]
\centering
\caption{Performance comparison with different adapters.}
\vspace{-2mm}
\resizebox{0.48\textwidth}{!}
{
\begin{tabular}{l|c|c|c|c|c}
\hline
Method & mIoU & $S_{\alpha}$ & $F^w_{\beta}$ & $mE_{\phi}$ & MAE \\
\hline
Baseline    & 0.751 & 0.866 & 0.812 & 0.924 & 0.029\\
w/o LoRA~\cite{hu2021lora}    & 0.768 & 0.872 & 0.816 & 0.921 & 0.028\\
w/o Adapter~\cite{houlsby2019parameter} & 0.774 & 0.875 & 0.822 & 0.924 & 0.028 \\
Full        & 0.789 & 0.884 & 0.838 & 0.933 & 0.023 \\
\hline
\end{tabular}
}
\label{table:abla_5}
\end{table} 
\section{Conclusion}
In this paper, we propose a novel feature learning framework named Dual-SAM for MAS.
The framework includes a dual structure with SAM's paradigm to enhance feature learning of marine images.
To instruct comprehensive underwater prior information, we propose a Multi-level Coupled Prompt (MCP) strategy.
In addition, we design a Dilated Fusion Attention Module (DFAM) and a Criss-Cross Connectivity Prediction ($C^3$P) to improve the localization perception of marine animals.
Extensive experiments show that our proposed method achieve state-of-the-art performances on five widely-used MAS datasets.

\small
\textbf{Acknowledgements.}
This work was supported in part by the National Natural Science Foundation of China (No.62101092), the Fundamental Research Funds for the Central Universities (No.DUT23YG232) and the Open Project Program of State Key Laboratory of Virtual Reality Technology and Systems, Beihang University (No.VRLAB2022C02).
\vspace*{-4mm} 
\clearpage
\clearpage
\setcounter{page}{1}
\maketitlesupplementary

\section{Introduction}
\label{sec:rationale}
In the main paper, we have provided quantitative comparisons with some existing methods as well as the ablation studies.
In this supplementary material, we first provide details of evaluation metrics.
Then, we compare our method with more methods.
Afterwards, we verify the transferability and zero-shot ability of our proposed method.
In addition, we further validate the effectiveness of MCP and PMS through more ablation results.
Finally, we present some visual results to show the effects of key modules.
\section{Evaluation Metrics}
In this section, we provide details of the five evaluation metrics used to assess compared models.
With these metrics, we can comprehensively and adequately demonstrate the superior performance of our model.

1) The Mean Intersection over Union (mIoU) is computed by first determining the Intersection over Union (IoU) for each individual class, and then averaging these values across all classes. It can be represented as:
\begin{equation}
IoU=\frac{|A \cap B|}{|A \cup B|}, mIoU=\frac{1}{C} \sum_{i=1}^{C}IoU_{i},
\end{equation}
where $A$ represents the predicted values for a certain class and $B$ represents the true values of that class.

2) The weighted F-measure ($F_\beta^w$) is determined by computing the \(F_\beta\) score for each class and then weighting each class's contribution according to its occurrence frequency in the dataset.  This metric can emphasize the performance on less-represented classes. It can be represented as:
\begin{equation}
        F_{\beta}=\frac{\left(1+\beta^{2}\right) \times \text { Precision }^{\omega} \times \text { Recall }^{\omega}}{\beta^{2} \times \text { Precision }^{\omega}+ \text { Recall }^{\omega}}
\end{equation}
where $Precision$ and $Recall$ are the precision and recall scores. $\beta$ is a parameter to trade-off the precision and recall. It is usually set to 0.3.

3) The structural similarity measure ($S_\alpha$)~\cite{fan2018enhanced} is a metric used to evaluate the structural similarity between two images. $S_\alpha$ aligns more closely with the human visual judgment of image quality.

4) The Mean Enhanced-Alignment Measure (\(mE_\phi\))~\cite{wang2004image} is a metric that merges local pixel information and overall image means into a single score. This metric effectively captures both the global statistics of the image and the nuances of local pixel alignments.

5) The Mean Absolute Error (MAE) quantifies the average of the absolute discrepancies between the prediction and the ground truth. It offers an overall assessment without considering class boundaries. Superior performance is reflected in lower MAE values. It can be represented as:
\begin{equation}
MAE(p, g)=\frac{1}{m} \sum_{i=1}^{m}\left|p_{i}-g_{i}\right|
\end{equation}
where $p$ is the prediction and $g$ is the ground truth. $m$ is the pixel number.

With the aforementioned five metrics, we can fully assess the overall completeness of mask predictions while ensuring the reliability of object boundaries.
Therefore, achieving optimal results across these five metrics can sufficiently demonstrate the effectiveness of our model.
\section{More Comparison Results}
In the main paper, we compare most recent methods.
Here, we present more comparison results corresponding to more methods.
As shown in Tab.~\ref{mask3k_res}, Tab.~\ref{ufo_res} and Tab.~\ref{USOD10k}, the experimental results fully demonstrate the effectiveness of our proposed method.
\begin{table*}[]
\centering
\renewcommand\arraystretch{1.1}
\setlength\tabcolsep{5.5pt}
\begin{tabular}{c|c|c|c|c|c|c|c|c|c|c}
\hline
&\multicolumn{5}{c|}{\textbf{MAS3K}}        &\multicolumn{5}{c}{\textbf{RMAS}} \\ \cline{2-11}
\textbf{Method}&\textbf{mIoU} & $\textbf{S}_\alpha$ & $\textbf{F}_\beta^w $&$ \textbf{mE}_\phi$ & \textbf{MAE}& \textbf{mIoU} & $\textbf{S}_\alpha$ & $\textbf{F}_\beta^w$ & $\textbf{mE}_\phi$ & \textbf{MAE} \\
\hline
 UNet++~\cite{zhou2018unet++}   &0.506&0.726&0.552&0.790&0.083&0.558&0.763&0.644&0.835&0.046\\
 BASNet~\cite{qin2019basnet}   &0.677&0.826&0.724&0.862&0.046&0.707&0.847&0.771&0.907&0.032\\
 PFANet~\cite{zhao2019pyramid}   &0.405&0.690&0.471&0.768&0.086&0.556&0.767&0.582&0.810&0.051\\
 SCRN~\cite{wu2019stacked}   &0.693&0.839&0.730&0.869&0.041&0.695&0.842&0.731&0.878&0.030\\
 U2Net~\cite{qin2020u2}   &0.654&0.812&0.711&0.851&0.047&0.676&0.830&0.762&0.904&0.029\\
 SINet~\cite{fan2020camouflaged} &0.658&0.820&0.725&0.884&0.039&0.684&0.835&0.780&0.908&0.025\\
 PFNet~\cite{mei2021camouflaged}  &0.695&0.839&0.746&0.890&0.039&0.694&0.843&0.771&0.922&0.026\\
 RankNet~\cite{lv2021simultaneously} &0.658&0.812&0.722&0.867&0.043&0.704&0.846&0.772&0.927&0.026\\
 C2FNet~\cite{sun2021context}   &0.717&0.851&0.761&0.894&0.038&0.721&0.858&0.788&0.923&0.026\\
 ECDNet~\cite{li2021marine}   &0.711&0.850&0.766&0.901&0.036&0.664&0.823&0.689&0.854&0.036\\
 OCENet~\cite{liu2022modeling}  &0.667&0.824&0.703&0.868&0.052&0.680&0.836&0.752&0.900&0.030\\
 ZoomNet~\cite{pang2022zoom}   &0.736&0.862&0.780&0.898&0.032&0.728&0.855&0.795&0.915&0.022\\
 MASNet~\cite{fu2023masnet} &0.742&0.864&0.788&0.906&0.032&\textcolor{blue}{\textbf{0.731}}&\textcolor{red}{\textbf{0.862}}&\textcolor{blue}{\textbf{0.801}}&0.920&0.024\\
    \hline
    SETR~\cite{zheng2021rethinking}   &0.715&0.855&0.789&0.917&0.030&0.654&0.818&0.747&0.933&0.028\\
    TransUNet~\cite{chen2021transunet}   &0.739&0.861&0.805&0.919&0.029&0.688&0.832&0.776&\textcolor{blue}{\textbf{0.941}}&0.025\\

    H2Former~\cite{he2023h2former}   &\textcolor{blue}{\textbf{0.748}}&0.865&\textcolor{blue}{\textbf{0.810}}&\textcolor{blue}{\textbf{0.925}}&\textcolor{blue}{\textbf{0.028}}&0.717&0.844&0.799&0.931&\textcolor{blue}{\textbf{0.023}}\\
    \hline
    SAM~\cite{kirillov2023segment}  &0.566&0.763&0.656&0.807&0.059&0.445&0.697&0.534&0.790&0.053\\
    SAM-Adapter\cite{chen2023sam}   &0.714&0.847&0.782&0.914&0.033&0.656&0.816&0.752&0.927&0.027\\
    SAM-DADF~\cite{lai2023detect}   &0.742&\textcolor{blue}{\textbf{0.866}}&0.806&0.925&0.028&0.686&0.833&0.780&0.926&0.024\\
    \hline
\textbf{Dual-SAM}   &\textcolor{red}{\textbf{0.789}}&\textcolor{red}{\textbf{0.884}}&\textcolor{red}{\textbf{0.838}}&\textcolor{red}{\textbf{0.933}}&\textcolor{red}{\textbf{0.023}}&\textcolor{red}{\textbf{0.735}}&\textcolor{blue}{\textbf{0.860}}&\textcolor{red}{\textbf{0.812}}&\textcolor{red}{\textbf{0.944}}&\textcolor{red}{\textbf{0.022}}\\
\hline
\end{tabular}
\vspace{-2mm}
\caption{Performance comparison on MAS3K and RMAS. The best and second results are in red and blue, respectively.}
\label{mask3k_res}
\vspace{-4mm}
\end{table*}
\begin{table*}[h]
\centering
\renewcommand\arraystretch{1.1}
\setlength\tabcolsep{5.5pt}
\begin{tabular}{c|c|c|c|c|c|c|c|c|c|c}
        \hline
        &\multicolumn{5}{c|}{\textbf{UFO120}}        &\multicolumn{5}{c}{\textbf{RUWI}} \\ \cline{2-11}
    \textbf{Method}&\textbf{mIoU} & $\textbf{S}_\alpha$ & $\textbf{F}_\beta^w$ & $\textbf{m}\textbf{E}_\phi$ & \textbf{MAE}& \textbf{mIoU} & $\textbf{S}_\alpha$ & $\textbf{F}_\beta^w$ & $\textbf{m}\textbf{E}_\phi$ & \textbf{MAE} \\
        \hline
        UNet++~\cite{zhou2018unet++}    &0.412&0.459&0.433&0.451&0.409&0.586&0.714&0.678&0.790&0.145\\
    BASNet~\cite{qin2019basnet}   &0.710&0.809&0.793&0.865&0.097&0.841&0.871&0.895&0.922&0.056\\
    PFANet~\cite{zhao2019pyramid}   &0.677&0.752&0.723&0.815&0.129&0.773&0.765&0.811&0.867&0.096\\
    SCRN~\cite{wu2019stacked}   &0.678&0.783&0.760&0.839&0.106&0.830&0.847&0.883&0.925&0.059\\
    U2Net~\cite{qin2020u2}   &0.680&0.792&0.709&0.811&0.134&0.841&0.873&0.861&0.786&0.074\\
    SINet~\cite{fan2020camouflaged}   &0.767&0.837&0.834&0.890&0.079&0.785&0.789&0.825&0.872&0.096\\
    PFNet~\cite{mei2021camouflaged}   &0.570&0.708&0.550&0.683&0.216&0.864&0.883&0.870&0.790&0.062\\
    RankNet~\cite{lv2021simultaneously}   &0.739&0.823&0.772&0.828&0.101&0.865&0.886&0.889&0.759&0.056\\
    C2FNet~\cite{sun2021context}   &0.747&0.826&0.806&0.878&0.083&0.840&0.830&0.883&0.924&0.060\\
    ECDNet~\cite{li2021marine}   &0.693&0.783&0.768&0.848&0.103&0.829&0.812&0.871&0.917&0.064\\
    OCENet~\cite{liu2022modeling}  &0.605&0.725&0.668&0.773&0.161&0.763&0.791&0.798&0.863&0.115\\
    ZoomNet~\cite{pang2022zoom}   &0.616&0.702&0.670&0.815&0.174&0.739&0.753&0.771&0.817&0.137\\
    MASNet~\cite{fu2023masnet} &0.754&0.827&0.820&0.879&0.083&0.865&0.880&0.913&0.944&0.047\\
    \hline

    SETR~\cite{zheng2021rethinking}   &0.711&0.811&0.796&0.871&0.089&0.832&0.864&0.895&0.924&0.055\\
    TransUNet~\cite{chen2021transunet}   &0.752&0.825&0.827&0.888&0.079&0.854&0.872&0.910&0.940&0.048\\

    H2Former~\cite{he2023h2former}   &\textcolor{blue}{\textbf{0.780}}&\textcolor{blue}{\textbf{0.844}}&\textcolor{blue}{\textbf{0.845}}&\textcolor{blue}{\textbf{0.901}}&\textcolor{blue}{\textbf{0.070}}&0.871&0.884&0.919&0.945&0.045\\
    \hline
    SAM~\cite{kirillov2023segment}   &0.681&0.768&0.745&0.827&0.121&0.849&0.855&0.907&0.929&0.057\\
    SAM-Adapter~\cite{chen2023sam}   &0.757&0.829&0.834&0.884&0.081&0.867&0.878&0.913&\textcolor{blue}{\textbf{0.946}}&0.046\\
    SAM-DADF~\cite{lai2023detect}   &0.768&0.841&0.836&0.893&0.073&\textcolor{blue}{\textbf{0.881}}&\textcolor{blue}{\textbf{0.889}}&\textcolor{blue}{\textbf{0.925}}&0.940&\textcolor{blue}{\textbf{0.044}}\\
    \hline
\textbf{Dual-SAM}   &\textcolor{red}{\textbf{0.810}}&\textcolor{red}{\textbf{0.856}}&\textcolor{red}{\textbf{0.864}}&\textcolor{red}{\textbf{0.914}}&\textcolor{red}{\textbf{0.064}}&\textcolor{red}{\textbf{0.904}}&\textcolor{red}{\textbf{0.903}}&\textcolor{red}{\textbf{0.939}}&\textcolor{red}{\textbf{0.959}}&\textcolor{red}{\textbf{0.035}}\\
\hline
\end{tabular}
\vspace{-2mm}
\caption{Performance comparison on UFO120 and RUWI. The best and second results are in red and blue, respectively.}
\label{ufo_res}
\vspace{-4mm}
\end{table*}
\begin{table}[]
\centering
\begin{tabular}{c|c|c|c|c}

\hline
        &\multicolumn{4}{c}{\textbf{USOD10k}}        \\ \cline{2-5}
\textbf{Method}& $\textbf{S}_\alpha$ & $\textbf{m}\textbf{E}_\phi$ & $\textbf{maxF}$ & \textbf{MAE}\\ \hline
Itti~\cite{itti1998model}   &.6112&.6670&.4676&.1798\\
RCRR~\cite{yuan2017reversion}&.6449&.6898&.5592&.1831\\
DF~\cite{qu2017rgbd}  &.6410&.7576&.5589&.1400\\
CPD~\cite{wu2019cascaded}  &.9076 & .9484 & .8991 & .0290 \\
DMRA~\cite{piao2019depth}                     & .8746                     & .9274 & .8682 & .0422 \\
SAMNet  ~\cite{zhao2019pyramid}                  & .8875                     & .9382 & .8739 & .0396 \\
PoolNet ~\cite{liu2019simple}                 & .9152                     & .9562 & .9105 & .0283 \\
BASNet~\cite{qin2019basnet}                   & .9075                     & .9378 & .8849 & .0352 \\
EGNet ~\cite{zhao2019egnet}                    & .9125                     & .9488 & .9040 & .0291 \\
FC-SOD ~\cite{zhang2020few}                  & .7036                     & .7004 & .6231 & .0852 \\
LDF ~\cite{wei2020label}                     & .9135                     & .9574 & .9173 & .0260 \\
F3Net ~\cite{wei2020f3net}                   & .9140                     & .9599 & .9171 & .0251 \\
PFPN ~\cite{wang2020progressive}                    & .9090                     & .9547 & .9055 & .0302 \\
MINet ~\cite{pang2020multi}                   & .9105                     & .9501 & .9072 & .0287 \\
DASNet ~\cite{zhao2020depth}                  & .9204                     & .9603 & .9212 & .0245 \\
JL-DCF \cite{fu2020jl}                  & .9062                     & .9485 & .8978 & .0300 \\
UCNet ~\cite{zhang2020uc}                   & .8997                     & .9463 & .8968 & .0301 \\
S2MA ~\cite{liu2021learning}                    & .8664                     & .9208 & .8530 & .0558 \\
BBSNet ~\cite{fan2020bbs}                  & .9061                     & .9512 & .9056 & .0337 \\
DANet ~\cite{zhao2020single}                   & .9006                     & .9449 & .8934 & .0279 \\
SGL-KRN ~\cite{xu2021locate}                 & .9214     & .9633 & \textcolor{blue}{\textbf{.9245}} & .0237 \\
DCF ~\cite{ji2021calibrated}                     & .9116                     & .9541 & .9045 & .0312 \\
SPNet ~\cite{zhou2021specificity}                   & .9075                     & .9554 & .9069 & .0280 \\
HAINet  ~\cite{li2021hierarchical}                 & .9123                     & .9552 & .9116 & .0279 \\
VST ~\cite{liu2021swin}                      & .9136                     & .9614 & .9108 & .0267\\
TriTransNet ~\cite{liu2021tritransnet}              & .7889                     & .8479 & .7501 & .0659 \\
CSNet ~\cite{cheng2021highly}                   & .8595                     & .9178 & .8462 & .0548 \\
D3Net ~\cite{fan2020rethinking}                   & .8931                     & .9413 & .8807 & .0374 \\
SVAM-Net ~\cite{islam2020svam}                & .7465                     &.7649  & .6451 & .0915\\
BTS-Net ~\cite{zhang2021bts}                  & .9093                     & .9542 & .9104 & .0291 \\
CDINet ~\cite{zhang2021cross}                  & .7049                     & .8644 & .7362 & .0904 \\
CTDNet  ~\cite{zhao2021complementary}                 & .9085                     & .9531 & .9073 & .0285   \\
MFNet ~\cite{piao2021mfnet}                   & .8425                     & .9146 & .8193 & .0512   \\
PFSNet ~\cite{ma2021pyramidal}                  & .8983                     & .9421 & .8966 & .0370 \\
PSGLoss ~\cite{yang2021progressive}                 & .8640                     & .9078 & .8508 & .0417 \\
TC-USOD~\cite{hong2023usod10k}                  & \textcolor{blue}{\textbf{.9215}}&\textcolor{blue}{\textbf{.9683}}&.9236&\textcolor{blue}{\textbf{.0201}}\\
\hline
SAM~\cite{kirillov2023segment}&.8543&.9095&.8812&.0380\\
SAM-Adapter~\cite{chen2023sam}&.8952&.9533&.9153&.0276\\
SAM-DADF~\cite{lai2023detect}&.9051&.9552&.9154&.0250\\\hline
\textbf{Dual-SAM}       &\textcolor{red}{\textbf{.9238}}    & \textcolor{red}{\textbf{.9684}} &\textcolor{red}{\textbf{.9311}}  &\textcolor{red}{\textbf{.0185}}\\ \hline
\end{tabular}
\vspace{-2mm}
\caption{Performance comparison on USOD10k. The best and second results are in red and blue, respectively.}
\label{USOD10k}
\end{table}
\section{Transferability and Zero-shot Ability}
In fact, our model can adapt to other complex tasks, such as saliency detection, camouflaged object detection and polyp segmentation.
To verify this fact, we conduct zero-shot and transferability testing on other datasets with large domain gaps, i.e., DUTS, COD10K and Kvasir.
As shown in Tab.~\ref{tab:other}, our method also achieves better results than other SAM-based methods and task-specific ones.
These results clearly verify the generalization of our method.
In addition, since we freeze SAM's encoder, it somewhat preserves the zero-shot ability.
As shown in Tab.~\ref{tab:other}, our method delivers comparable results with SAM, showing an expressive zero-shot ability.
\begin{table}[htp]
\vspace{-2mm}
\centering
\resizebox{0.48\textwidth}{!}
{
\begin{tabular}{c|c|c|c|c|c|ccccc}
\hline
&\multicolumn{2}{c|}{\textbf{DUTS (SOD)}} &\multicolumn{2}{c|}{\textbf{COD10K (COD)}}&\multicolumn{2}{c}{\textbf{Kvasir (Medical)}} \\ \cline{2-7}
\textbf{Method} & $\textbf{F}_\beta^w$ & \textbf{MAE} & $\textbf{F}_\beta^w$ & \textbf{MAE}& $\textbf{F}_\beta^w$ & \textbf{MAE} \\
\hline
VST        &0.828&0.037&-----&-----&-----&-----\\
PFNet      &-----&-----&0.660&0.040&-----&-----\\
FAPNet     &-----&-----&-----&-----&0.894&0.027\\
\hline
SAM        &0.764&0.058&0.633&0.050&0.769&0.062\\
SAM-Adapter&0.878&0.029&0.801&0.025&0.876&0.029\\
Ours (zero-shot)&0.783&0.048&0.677&0.044&0.696&0.082\\
Ours       &\textbf{0.885}&\textbf{0.025}&\textbf{0.889}&\textbf{0.012}&\textbf{0.909}&\textbf{0.025}\\
\hline
\end{tabular}
}
\vspace{-2mm}
\caption{Performance comparison on other complex tasks.}
\label{tab:other}
\end{table}
\section{More Ablation Results on MCP and PMS}
Experiments are conducted on MAS3K~\cite{li2020mas3k} for its challenging and high-quality annotations.

\textbf{Effects of MCP.}
For MCP, we first enhance the features through a self-attention mechanism, and then integrate the features extracted from SAM by using a cross-attention mechanism.
In Tab.~\ref{table:mcp}, we compare the effectiveness of these internal components of MCP.
In the second and third rows, we list the results of using the self-attention mechanism ($S_{only}$MCP) and the cross-attention mechanism ($C_{only}$MCP), respectively.
Compared with the whole MCP structure in the last row, it indicates that both mechanisms have a positive effect.
%

\textbf{Effects of PMS.}
In Tab.~\ref{table:pms}, we compare the impact of using mutual supervision at different decoder layers.
``1 PMS'' refers to the incorporation of the mutual supervision module in the first layer of the decoder, and the other definitions follow similarly.
As the number of layers increases, the performance gradually improves.
We can observe that mutual supervision has a positive effect.
With mutual supervision between the two branches, the objects' details are adequately complemented.
\begin{table}[h!]
\centering
\small
\begin{tabular}{l|c|c|c|c|c}
\hline
Method &\textbf{mIoU} & $\textbf{S}_\alpha$ & $\textbf{F}_\beta^w$ & $\textbf{m}\textbf{E}_\phi$ & \textbf{MAE} \\
\hline
no MCP & 0.778 & 0.877 & 0.825 & 0.929 & 0.026 \\
$S_{only}$ MCP &0.779&0.878&0.828&0.931&0.026\\
$C_{only}$ MCP &0.783&0.879&0.832&0.931&0.025\\
MCP   & 0.789 & 0.884 & 0.838 & 0.933 & 0.023 \\
\hline
\end{tabular}
\caption{Performance comparisons of MCP.}
\label{table:mcp}
\end{table}
\begin{table}[h!]
\centering
\small
\begin{tabular}{l|c|c|c|c|c}
\hline
Method &\textbf{mIoU} & $\textbf{S}_\alpha$ & $\textbf{F}_\beta^w$ & $\textbf{m}\textbf{E}_\phi$ & \textbf{MAE} \\
\hline
no PMS & 0.771 & 0.874 & 0.820 & 0.923 & 0.029 \\
1 PMS & 0.776 & 0.876 & 0.823 & 0.926 & 0.027 \\
2 PMS & 0.779 & 0.878 & 0.827 & 0.927 & 0.026 \\
3 PMS & 0.783 & 0.880 & 0.830 & 0.932 & 0.025 \\
4 PMS   & 0.789 & 0.884 & 0.838 & 0.933 & 0.023 \\
\hline
\end{tabular}
\caption{Performance comparisons with different layers of PMS.}
\label{table:pms}
\end{table}
\section{More Visual Results}
In the main paper, we have already presented a visual comparison of typical methods.
In this supplementary material, we provide more visual results to verify the effects of our proposed key modules.
%

\begin{figure}
\renewcommand\thesubfigure{\alph{subfigure}}

    \centering
    \begin{subfigure}{0.09\textwidth}
        \includegraphics[width=\linewidth]{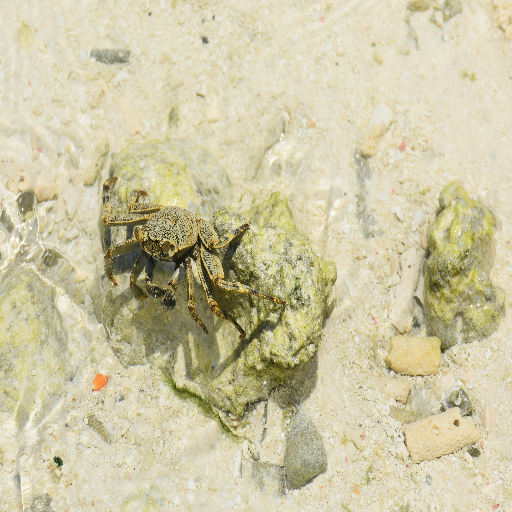}
        \caption{Image}
    \end{subfigure}
    \begin{subfigure}{0.09\textwidth}
        \includegraphics[width=\linewidth]{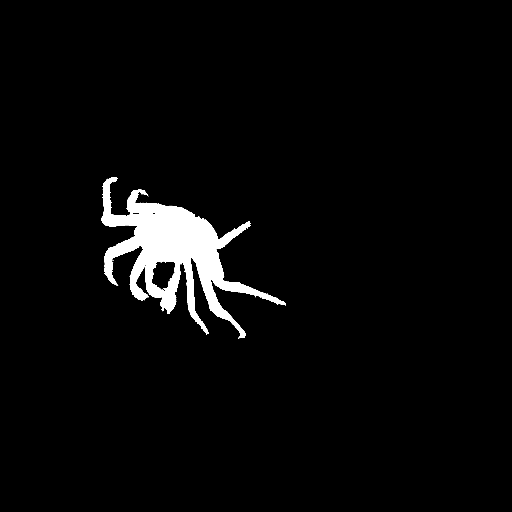}
        \caption{GT}
    \end{subfigure}
    \begin{subfigure}{0.09\textwidth}
        \includegraphics[width=\linewidth]{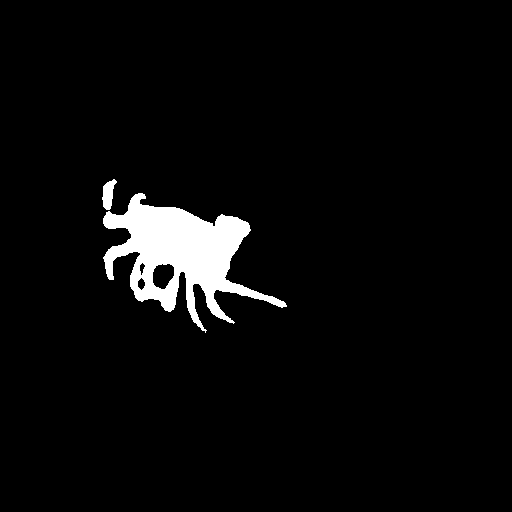}
        \caption{$C^3$P}
    \end{subfigure}
    \begin{subfigure}{0.09\textwidth}
        \includegraphics[width=\linewidth]{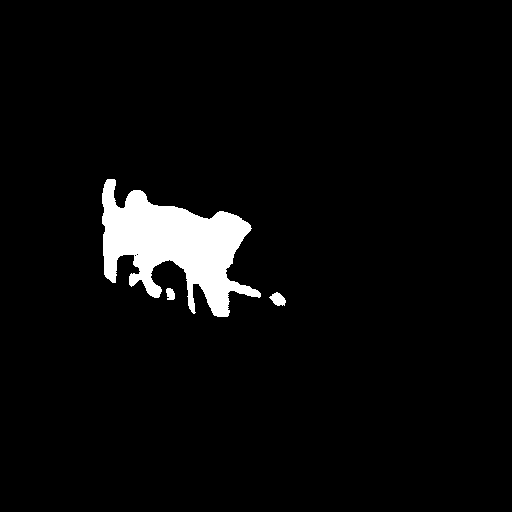}
        \caption{Binary}
    \end{subfigure}
    \begin{subfigure}{0.09\textwidth}
        \includegraphics[width=\linewidth]{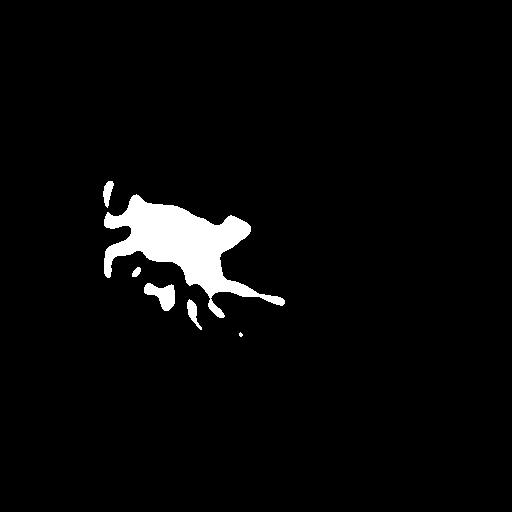}
        \caption{Nearby}
    \end{subfigure}
    \caption{Visualizing the effect of our $C^3$P module.}
    \label{ab1}
\end{figure}

\begin{figure}
\renewcommand\thesubfigure{\alph{subfigure}}

    \centering
    \begin{subfigure}{0.09\textwidth}
        \includegraphics[width=\linewidth]{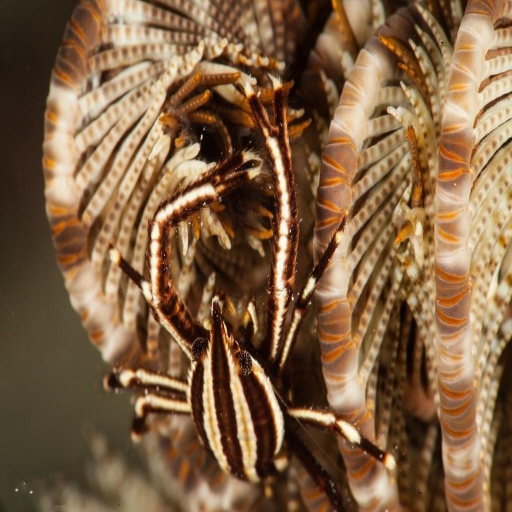}
        \caption{Image}
    \end{subfigure}
    \begin{subfigure}{0.09\textwidth}
        \includegraphics[width=\linewidth]{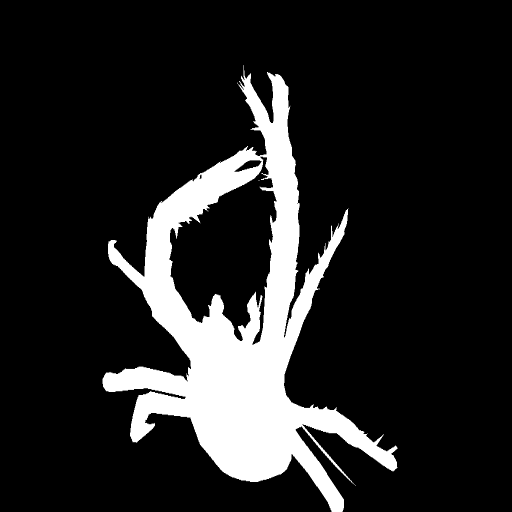}
        \caption{GT}
    \end{subfigure}
    \begin{subfigure}{0.09\textwidth}
        \includegraphics[width=\linewidth]{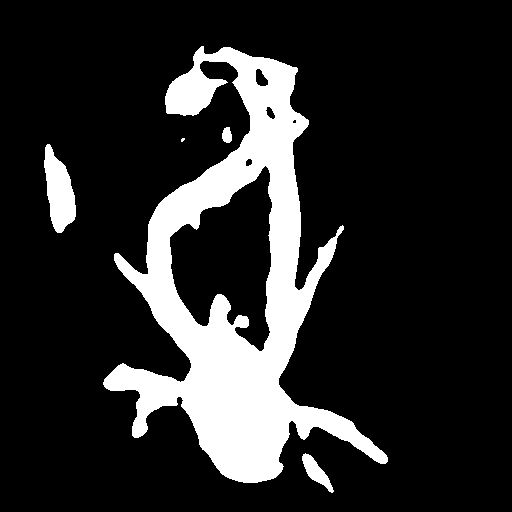}
        \caption{PMS}
    \end{subfigure}
    \begin{subfigure}{0.09\textwidth}
        \includegraphics[width=\linewidth]{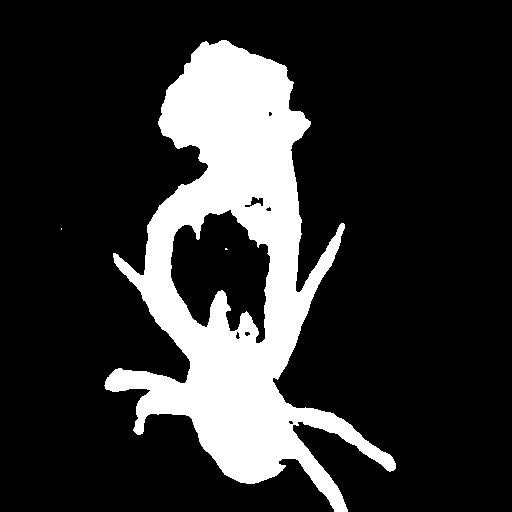}
        \caption{Single}
    \end{subfigure}
    \begin{subfigure}{0.09\textwidth}
        \includegraphics[width=\linewidth]{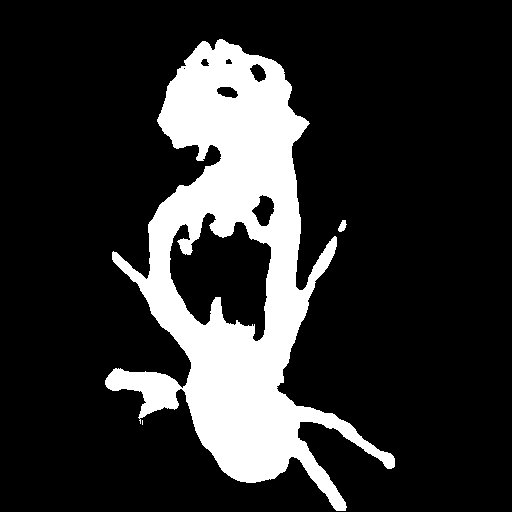}
        \caption{Dual}
    \end{subfigure}
    \caption{Visualizing the effect of our PMS module.}
    \label{ab2}
\end{figure}

\begin{figure}
\renewcommand\thesubfigure{\alph{subfigure}}
    \centering
    \begin{subfigure}{0.09\textwidth}
        \includegraphics[width=\linewidth]{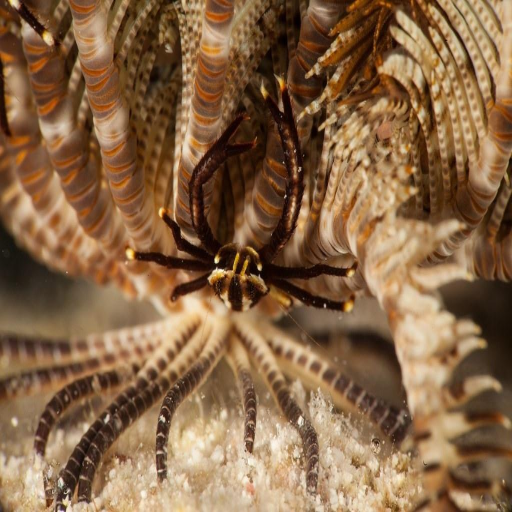}
        \caption{Image}
    \end{subfigure}
    \begin{subfigure}{0.09\textwidth}
        \includegraphics[width=\linewidth]{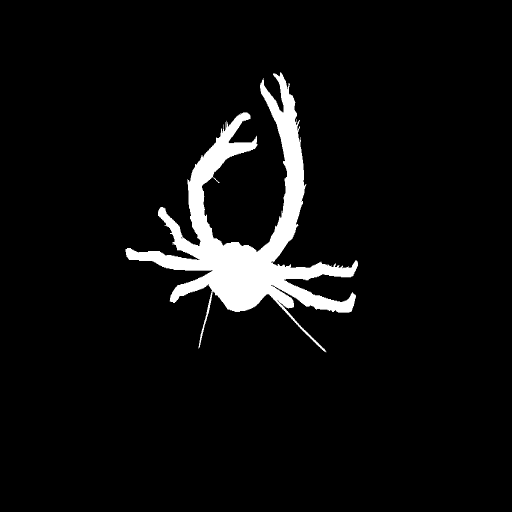}
        \caption{GT}
    \end{subfigure}
    \begin{subfigure}{0.09\textwidth}
        \includegraphics[width=\linewidth]{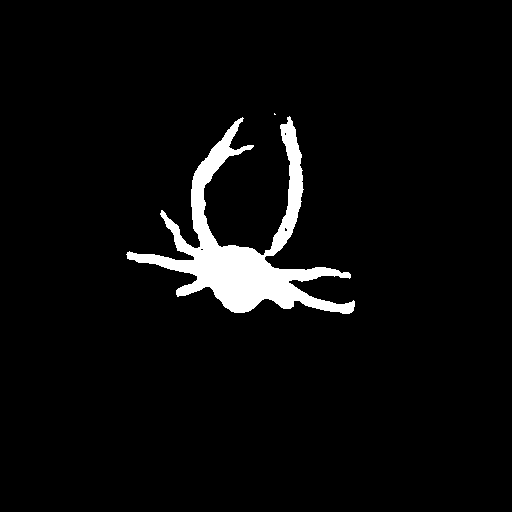}
        \caption{MCP}
    \end{subfigure}
    \begin{subfigure}{0.09\textwidth}
        \includegraphics[width=\linewidth]{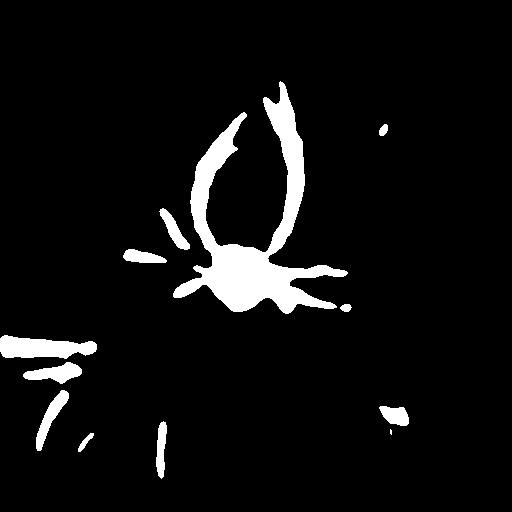}
        \caption{no MCP}
    \end{subfigure}
    \caption{Visualizing the effect of our MCP module.}
    \label{ab3}
\end{figure}

\begin{figure}
\renewcommand\thesubfigure{\alph{subfigure}}

    \centering
    \begin{subfigure}{0.09\textwidth}
        \includegraphics[width=\linewidth]{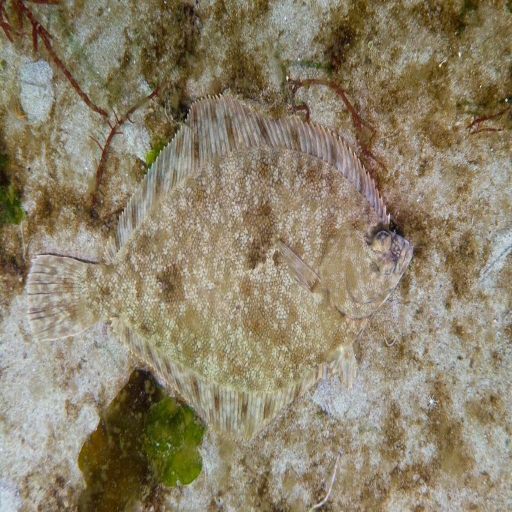}
        \caption{Image}
    \end{subfigure}
    \begin{subfigure}{0.09\textwidth}
        \includegraphics[width=\linewidth]{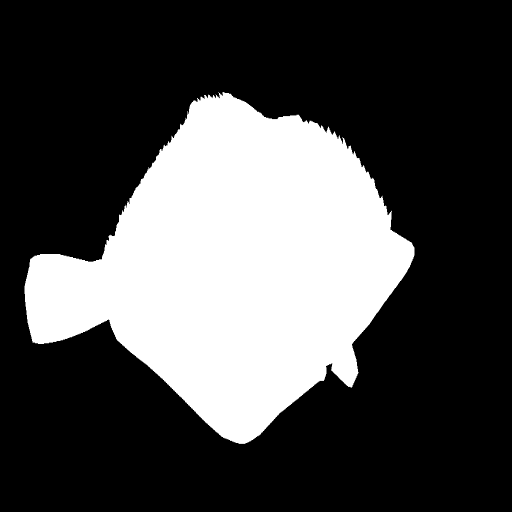}
        \caption{GT}
    \end{subfigure}
    \begin{subfigure}{0.09\textwidth}
        \includegraphics[width=\linewidth]{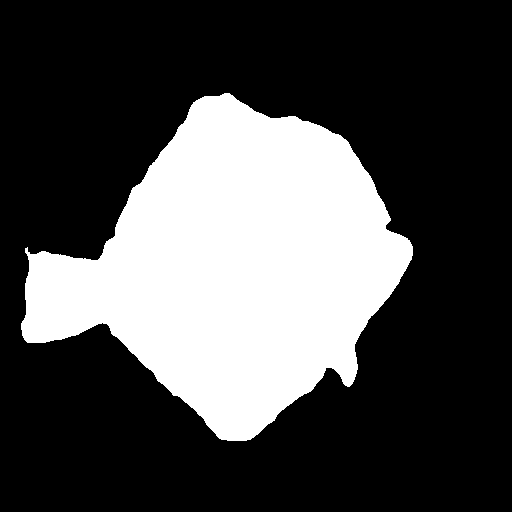}
        \caption{DFAM}
    \end{subfigure}
    \begin{subfigure}{0.09\textwidth}
        \includegraphics[width=\linewidth]{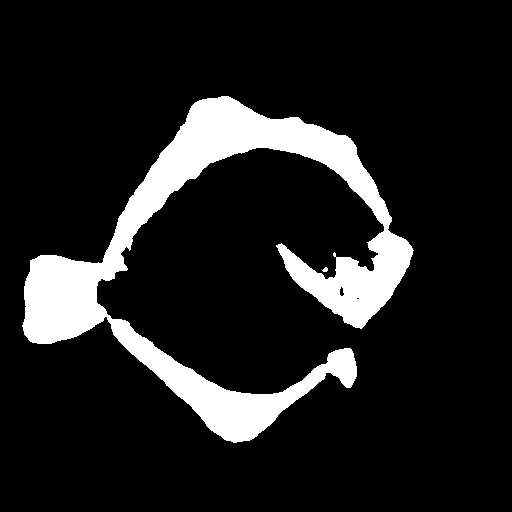}
        \caption{no DFAM}
    \end{subfigure}
    \caption{Visualizing the effect of our DFAM module.}
    \label{ab4}
\end{figure}

\begin{figure}
\renewcommand\thesubfigure{\alph{subfigure}}
    \centering
    \begin{subfigure}{0.09\textwidth}
        \includegraphics[width=\linewidth]{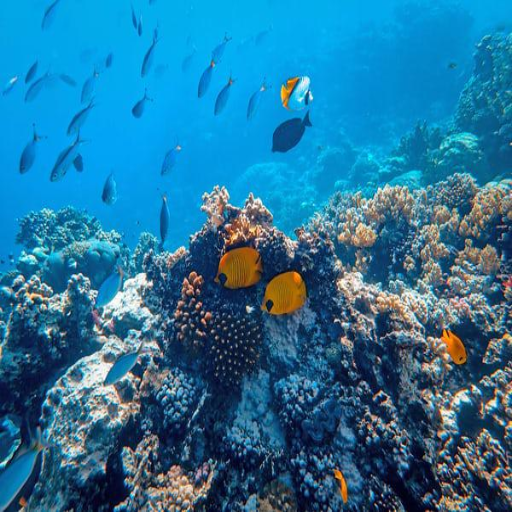}
        \caption{Image}
    \end{subfigure}
    \begin{subfigure}{0.09\textwidth}
        \includegraphics[width=\linewidth]{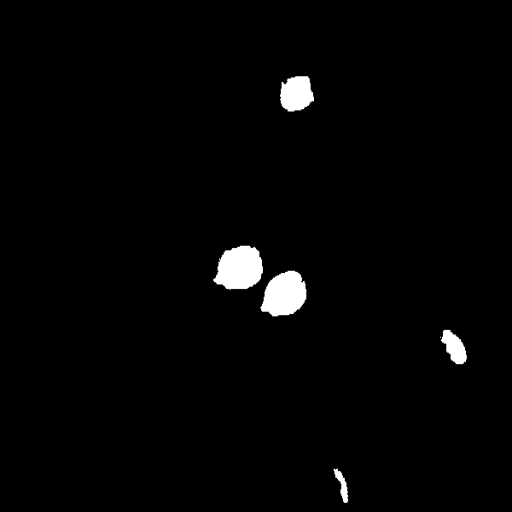}
        \caption{GT}
    \end{subfigure}
    \begin{subfigure}{0.09\textwidth}
        \includegraphics[width=\linewidth]{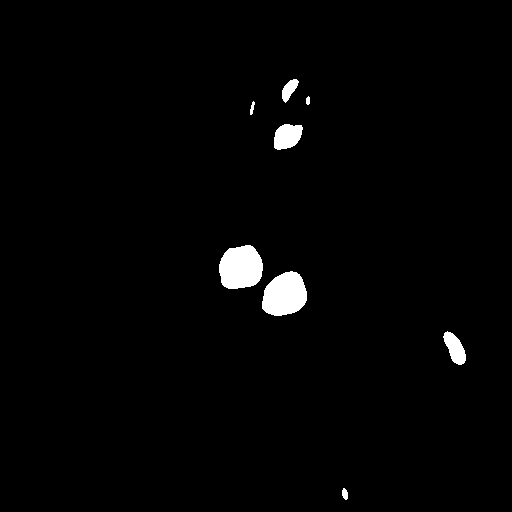}
        \caption{Ours}
    \end{subfigure}
    \begin{subfigure}{0.09\textwidth}
        \includegraphics[width=\linewidth]{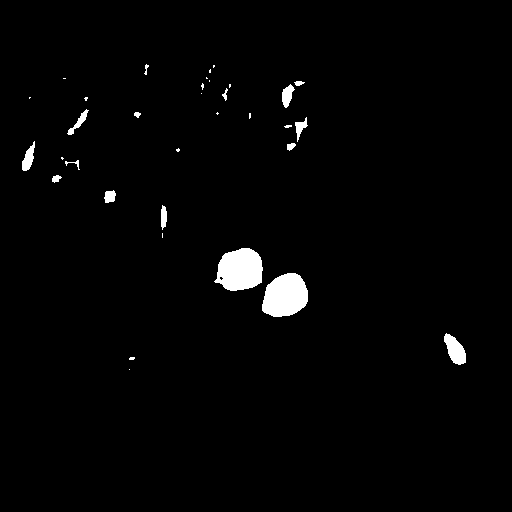}
        \caption{LoRA}
    \end{subfigure}
    \begin{subfigure}{0.09\textwidth}
        \includegraphics[width=\linewidth]{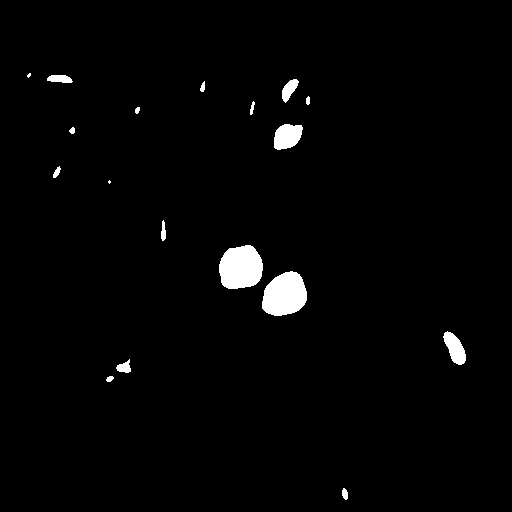}
        \caption{Adapter}
    \end{subfigure}
    \caption{Visualizing the effect of different adapter mechanisms.}
    \label{ab5}
\end{figure}

\textbf{Visual Results with Key Modules.}
In Fig.~\ref{ab1}, we show the visual effect of our $C^3$P module.
One can observe that our $C^3$P module helps to obtain a better overall shape of underwater targets.
The binary cross-entropy loss and nearby connectivity prediction are not good at predicting the animal boundaries
In Fig.~\ref{ab2}, we show the visual effect of our PMS module.
By employing dual branches for mutual supervision, the segmentation maps have comprehensive information, effectively removing redundant information.
In Fig.~\ref{ab3}, we show the visual effect of our MCP module.
With the multi-level coupled guidance, SAM has gained enhanced representational capabilities for animals and suppressed the cluttered backgrounds.
In Fig.~\ref{ab4}, we show the visual effect of our DFAM module.
We integrate the features extracted from both the encoder and decoder through the DFAM module, and select more important feature channels.
The design can adaptively aggregate more contextual information and significantly improve the segmentation results.
In Fig.~\ref{ab5}, we show the visual effect of our adapter mechanism.
One can observe that our method effectively injects underwater domain information into the SAM backbone.
Furthermore, the use of our dual adapter mechanisms continues to have a positive impact on the performance.

\begin{figure*}
\renewcommand\thesubfigure{\alph{subfigure}}
    \centering
    \begin{subfigure}{0.1\textwidth}
        \includegraphics[width=\linewidth]{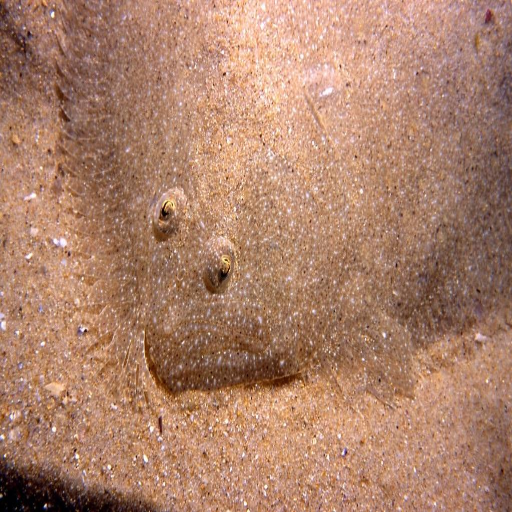}
        \caption{Image}
    \end{subfigure}
    \begin{subfigure}{0.1\textwidth}
        \includegraphics[width=\linewidth]{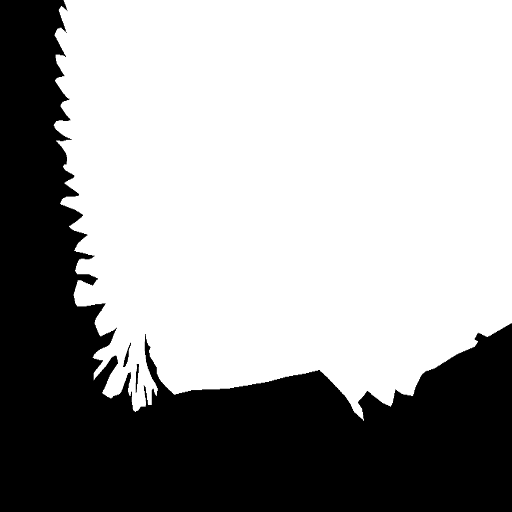}
        \caption{GT}
    \end{subfigure}
    \begin{subfigure}{0.1\textwidth}
        \includegraphics[width=\linewidth]{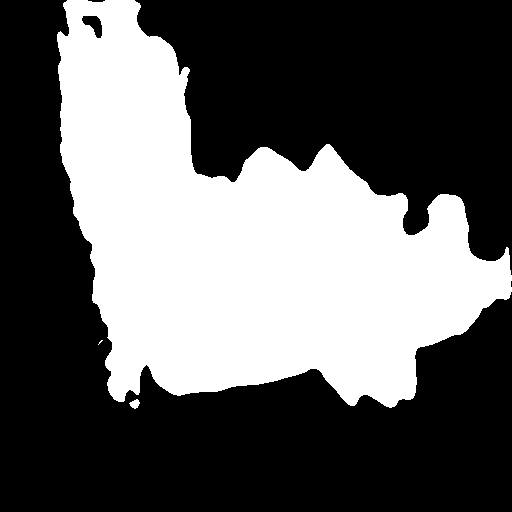}
        \caption{Dual-SAM}
    \end{subfigure}
    \begin{subfigure}{0.1\textwidth}
        \includegraphics[width=\linewidth]{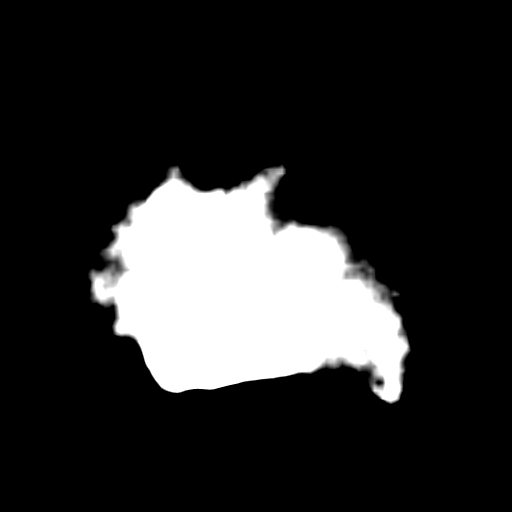}
        \caption{SETR}
    \end{subfigure}
    \begin{subfigure}{0.1\textwidth}
        \includegraphics[width=\linewidth]{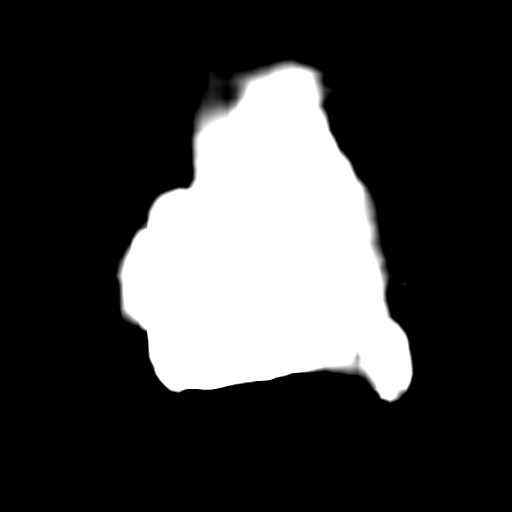}
        \caption{TransUNet}
    \end{subfigure}
    \begin{subfigure}{0.1\textwidth}
        \includegraphics[width=\linewidth]{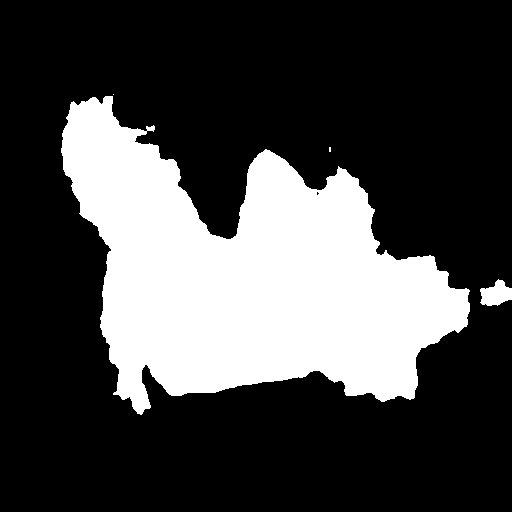}
        \caption{H2Former}
    \end{subfigure}
    \begin{subfigure}{0.1\textwidth}
        \includegraphics[width=\linewidth]{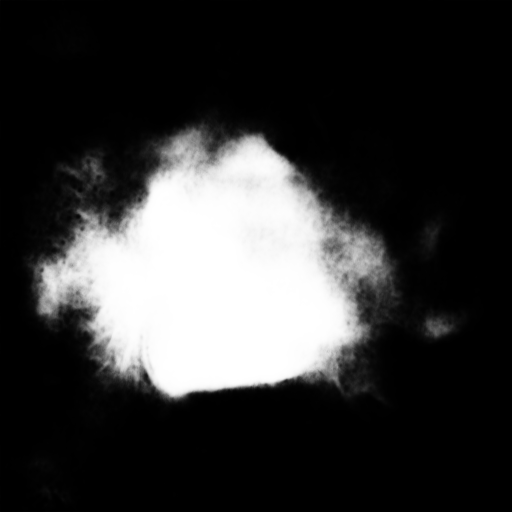}
        \caption{SAM}
    \end{subfigure}
    \begin{subfigure}{0.1\textwidth}
        \includegraphics[width=\linewidth]{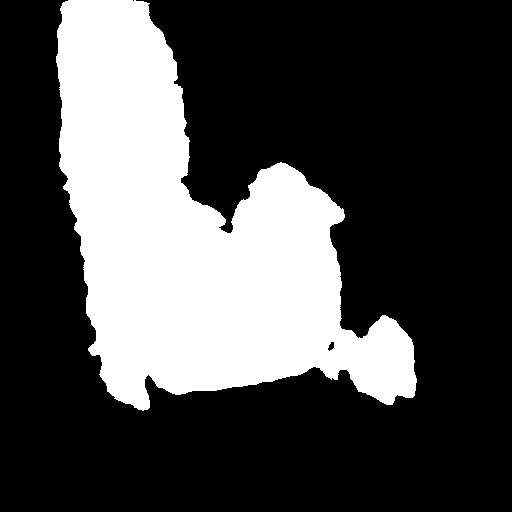}
        \caption{Adapter}
    \end{subfigure}
    \begin{subfigure}{0.1\textwidth}
        \includegraphics[width=\linewidth]{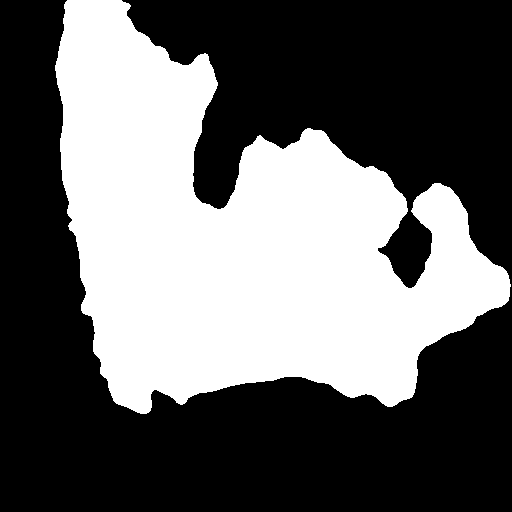}
        \caption{DADF}
    \end{subfigure}
    \caption{Visualizing failure segmentation cases.}
    \label{ab6}
\end{figure*}

\textbf{Visualization of Failed Results.}
In Fig.~\ref{ab6}, we present some failure cases.
Due to the similarity between the animal and its environment, it is challenging for our model to capture it accurately.
However, other existing methods also result in significant segmentation errors.
Therefore, distinguishing such organisms has become a focus of our further efforts.

{
    \small
    \bibliographystyle{ieeenat_fullname}
    \bibliography{main}

\begin{thebibliography}{89}
\providecommand{\natexlab}[1]{#1}
\providecommand{\url}[1]{\texttt{#1}}
\expandafter\ifx\csname urlstyle\endcsname\relax
  \providecommand{\doi}[1]{doi: #1}\else
  \providecommand{\doi}{doi: \begingroup \urlstyle{rm}\Url}\fi

\bibitem[Bay et~al.(2008)Bay, Ess, Tuytelaars, and Van~Gool]{bay2008speeded}
Herbert Bay, Andreas Ess, Tinne Tuytelaars, and Luc Van~Gool.
\newblock Speeded-up robust features (surf).
\newblock \emph{CVIU}, 110\penalty0 (3):\penalty0 346--359, 2008.

\bibitem[Chen et~al.(2021)Chen, Lu, Yu, Luo, Adeli, Wang, Lu, Yuille, and
  Zhou]{chen2021transunet}
Jieneng Chen, Yongyi Lu, Qihang Yu, Xiangde Luo, Ehsan Adeli, Yan Wang, Le Lu,
  Alan~L Yuille, and Yuyin Zhou.
\newblock Transunet: Transformers make strong encoders for medical image
  segmentation.
\newblock \emph{arXiv}, 2021.

\bibitem[Chen et~al.(2023{\natexlab{a}})Chen, Liu, Chen, Zhang, Li, Zou, and
  Shi]{chen2023rsprompter}
Keyan Chen, Chenyang Liu, Hao Chen, Haotian Zhang, Wenyuan Li, Zhengxia Zou,
  and Zhenwei Shi.
\newblock Rsprompter: Learning to prompt for remote sensing instance
  segmentation based on visual foundation model.
\newblock \emph{arXiv}, 2023{\natexlab{a}}.

\bibitem[Chen et~al.(2017)Chen, Papandreou, Kokkinos, Murphy, and
  Yuille]{chen2017deeplab}
Liang-Chieh Chen, George Papandreou, Iasonas Kokkinos, Kevin Murphy, and Alan~L
  Yuille.
\newblock Deeplab: Semantic image segmentation with deep convolutional nets,
  atrous convolution, and fully connected crfs.
\newblock \emph{IEEE TPAMI}, 40\penalty0 (4):\penalty0 834--848, 2017.

\bibitem[Chen et~al.(2022)Chen, Fu, Huang, Cheng, and Ding]{chen2022robust}
Ruizhe Chen, Zhenqi Fu, Yue Huang, En Cheng, and Xinghao Ding.
\newblock A robust object segmentation network for underwater scenes.
\newblock In \emph{ICASSP}, pages 2629--2633. IEEE, 2022.

\bibitem[Chen et~al.(2023{\natexlab{b}})Chen, Zhu, Ding, Cao, Zhang, Wang, Li,
  Sun, Mao, and Zang]{chen2023sam}
Tianrun Chen, Lanyun Zhu, Chaotao Ding, Runlong Cao, Shangzhan Zhang, Yan Wang,
  Zejian Li, Lingyun Sun, Papa Mao, and Ying Zang.
\newblock Sam fails to segment anything?--sam-adapter: Adapting sam in
  underperformed scenes: Camouflage, shadow, and more.
\newblock \emph{arXiv}, 2023{\natexlab{b}}.

\bibitem[Cheng et~al.(2021)Cheng, Gao, Borji, Tan, Lin, and
  Wang]{cheng2021highly}
Ming-Ming Cheng, Shang-Hua Gao, Ali Borji, Yong-Qiang Tan, Zheng Lin, and Meng
  Wang.
\newblock A highly efficient model to study the semantics of salient object
  detection.
\newblock \emph{PAMI}, 44\penalty0 (11):\penalty0 8006--8021, 2021.

\bibitem[Dosovitskiy et~al.(2020)Dosovitskiy, Beyer, Kolesnikov, Weissenborn,
  Zhai, Unterthiner, Dehghani, Minderer, Heigold, Gelly,
  et~al.]{dosovitskiy2020image}
Alexey Dosovitskiy, Lucas Beyer, Alexander Kolesnikov, Dirk Weissenborn,
  Xiaohua Zhai, Thomas Unterthiner, Mostafa Dehghani, Matthias Minderer, Georg
  Heigold, Sylvain Gelly, et~al.
\newblock An image is worth 16x16 words: Transformers for image recognition at
  scale.
\newblock \emph{arXiv}, 2020.

\bibitem[Drews-Jr et~al.(2021)Drews-Jr, Souza, Maurell, Protas, and
  C.~Botelho]{drews2021underwater}
Paulo Drews-Jr, Isadora~de Souza, Igor~P Maurell, Eglen~V Protas, and Silvia~S
  C.~Botelho.
\newblock Underwater image segmentation in the wild using deep learning.
\newblock \emph{Journal of the Brazilian Computer Society}, 27:\penalty0 1--14,
  2021.

\bibitem[Fan et~al.(2018)Fan, Gong, Cao, Ren, Cheng, and
  Borji]{fan2018enhanced}
Deng-Ping Fan, Cheng Gong, Yang Cao, Bo Ren, Ming-Ming Cheng, and Ali Borji.
\newblock Enhanced-alignment measure for binary foreground map evaluation.
\newblock \emph{arXiv preprint arXiv:1805.10421}, 2018.

\bibitem[Fan et~al.(2020{\natexlab{a}})Fan, Ji, Sun, Cheng, Shen, and
  Shao]{fan2020camouflaged}
Deng-Ping Fan, Ge-Peng Ji, Guolei Sun, Ming-Ming Cheng, Jianbing Shen, and Ling
  Shao.
\newblock Camouflaged object detection.
\newblock In \emph{CVPR}, pages 2777--2787, 2020{\natexlab{a}}.

\bibitem[Fan et~al.(2020{\natexlab{b}})Fan, Lin, Zhang, Zhu, and
  Cheng]{fan2020rethinking}
Deng-Ping Fan, Zheng Lin, Zhao Zhang, Menglong Zhu, and Ming-Ming Cheng.
\newblock Rethinking rgb-d salient object detection: Models, data sets, and
  large-scale benchmarks.
\newblock \emph{TNNLS}, 32\penalty0 (5):\penalty0 2075--2089,
  2020{\natexlab{b}}.

\bibitem[Fan et~al.(2020{\natexlab{c}})Fan, Zhai, Borji, Yang, and
  Shao]{fan2020bbs}
Deng-Ping Fan, Yingjie Zhai, Ali Borji, Jufeng Yang, and Ling Shao.
\newblock Bbs-net: Rgb-d salient object detection with a bifurcated backbone
  strategy network.
\newblock In \emph{ECCV}, pages 275--292. Springer, 2020{\natexlab{c}}.

\bibitem[Fu et~al.(2020)Fu, Fan, Ji, and Zhao]{fu2020jl}
Keren Fu, Deng-Ping Fan, Ge-Peng Ji, and Qijun Zhao.
\newblock Jl-dcf: Joint learning and densely-cooperative fusion framework for
  rgb-d salient object detection.
\newblock In \emph{CVPR}, pages 3052--3062, 2020.

\bibitem[Fu et~al.(2023)Fu, Chen, Huang, Cheng, Ding, and Ma]{fu2023masnet}
Zhenqi Fu, Ruizhe Chen, Yue Huang, En Cheng, Xinghao Ding, and Kai-Kuang Ma.
\newblock Masnet: A robust deep marine animal segmentation network.
\newblock \emph{IEEE Journal of Oceanic Engineering}, 2023.

\bibitem[Gao et~al.(2023)Gao, Xia, Hu, and Gao]{gao2023desam}
Yifan Gao, Wei Xia, Dingdu Hu, and Xin Gao.
\newblock Desam: Decoupling segment anything model for generalizable medical
  image segmentation.
\newblock \emph{arXiv}, 2023.

\bibitem[He et~al.(2023)He, Wang, Li, Du, Xia, and Fu]{he2023h2former}
Along He, Kai Wang, Tao Li, Chengkun Du, Shuang Xia, and Huazhu Fu.
\newblock H2former: An efficient hierarchical hybrid transformer for medical
  image segmentation.
\newblock \emph{TMI}, 2023.

\bibitem[He et~al.(2016)He, Zhang, Ren, and Sun]{he2016deep}
Kaiming He, Xiangyu Zhang, Shaoqing Ren, and Jian Sun.
\newblock Deep residual learning for image recognition.
\newblock In \emph{CVPR}, pages 770--778, 2016.

\bibitem[Hendrycks and Gimpel(2016)]{hendrycks2016gaussian}
Dan Hendrycks and Kevin Gimpel.
\newblock Gaussian error linear units (gelus).
\newblock \emph{arXiv preprint arXiv:1606.08415}, 2016.

\bibitem[Hong et~al.(2023)Hong, Wang, Zhang, and Zhao]{hong2023usod10k}
Lin Hong, Xin Wang, Gan Zhang, and Ming Zhao.
\newblock Usod10k: a new benchmark dataset for underwater salient object
  detection.
\newblock \emph{TIP}, 2023.

\bibitem[Houlsby et~al.(2019)Houlsby, Giurgiu, Jastrzebski, Morrone,
  De~Laroussilhe, Gesmundo, Attariyan, and Gelly]{houlsby2019parameter}
Neil Houlsby, Andrei Giurgiu, Stanislaw Jastrzebski, Bruna Morrone, Quentin
  De~Laroussilhe, Andrea Gesmundo, Mona Attariyan, and Sylvain Gelly.
\newblock Parameter-efficient transfer learning for nlp.
\newblock In \emph{ICML}, pages 2790--2799. PMLR, 2019.

\bibitem[Hu et~al.(2021)Hu, Shen, Wallis, Allen-Zhu, Li, Wang, Wang, and
  Chen]{hu2021lora}
Edward~J Hu, Yelong Shen, Phillip Wallis, Zeyuan Allen-Zhu, Yuanzhi Li, Shean
  Wang, Lu Wang, and Weizhu Chen.
\newblock Lora: Low-rank adaptation of large language models.
\newblock \emph{arXiv}, 2021.

\bibitem[Huang et~al.(2017)Huang, Liu, Van Der~Maaten, and
  Weinberger]{huang2017densely}
Gao Huang, Zhuang Liu, Laurens Van Der~Maaten, and Kilian~Q Weinberger.
\newblock Densely connected convolutional networks.
\newblock In \emph{CVPR}, pages 4700--4708, 2017.

\bibitem[Islam et~al.(2020{\natexlab{a}})Islam, Luo, and
  Sattar]{islam2020simultaneous}
Md~Jahidul Islam, Peigen Luo, and Junaed Sattar.
\newblock Simultaneous enhancement and super-resolution of underwater imagery
  for improved visual perception.
\newblock \emph{arXiv}, 2020{\natexlab{a}}.

\bibitem[Islam et~al.(2020{\natexlab{b}})Islam, Wang, and
  Sattar]{islam2020svam}
Md~Jahidul Islam, Ruobing Wang, and Junaed Sattar.
\newblock Svam: saliency-guided visual attention modeling by autonomous
  underwater robots.
\newblock \emph{arXiv}, 2020{\natexlab{b}}.

\bibitem[Itti et~al.(1998)Itti, Koch, and Niebur]{itti1998model}
Laurent Itti, Christof Koch, and Ernst Niebur.
\newblock A model of saliency-based visual attention for rapid scene analysis.
\newblock \emph{PAMI}, 20\penalty0 (11):\penalty0 1254--1259, 1998.

\bibitem[Ji et~al.(2021)Ji, Li, Yu, Zhang, Piao, Yao, Bi, Ma, Zheng, Lu,
  et~al.]{ji2021calibrated}
Wei Ji, Jingjing Li, Shuang Yu, Miao Zhang, Yongri Piao, Shunyu Yao, Qi Bi, Kai
  Ma, Yefeng Zheng, Huchuan Lu, et~al.
\newblock Calibrated rgb-d salient object detection.
\newblock In \emph{CVPR}, pages 9471--9481, 2021.

\bibitem[Jin et~al.(2023)Jin, Chen, Chen, Xu, and Feng]{jin2023let}
Zheyan Jin, Shiqi Chen, Yueting Chen, Zhihai Xu, and Huajun Feng.
\newblock Let segment anything help image dehaze.
\newblock \emph{arXiv}, 2023.

\bibitem[Kampffmeyer et~al.(2018)Kampffmeyer, Dong, Liang, Zhang, and
  Xing]{kampffmeyer2018connnet}
Michael Kampffmeyer, Nanqing Dong, Xiaodan Liang, Yujia Zhang, and Eric~P Xing.
\newblock Connnet: A long-range relation-aware pixel-connectivity network for
  salient segmentation.
\newblock \emph{TIP}, 28\penalty0 (5):\penalty0 2518--2529, 2018.

\bibitem[Kirillov et~al.(2023)Kirillov, Mintun, Ravi, Mao, Rolland, Gustafson,
  Xiao, Whitehead, Berg, Lo, et~al.]{kirillov2023segment}
Alexander Kirillov, Eric Mintun, Nikhila Ravi, Hanzi Mao, Chloe Rolland, Laura
  Gustafson, Tete Xiao, Spencer Whitehead, Alexander~C Berg, Wan-Yen Lo, et~al.
\newblock Segment anything.
\newblock \emph{arXiv}, 2023.

\bibitem[Lai et~al.(2023)Lai, Luo, and Yu]{lai2023detect}
Yingxin Lai, Zhiming Luo, and Zitong Yu.
\newblock Detect any deepfakes: Segment anything meets face forgery detection
  and localization.
\newblock \emph{arXiv}, 2023.

\bibitem[Lane et~al.(1998)Lane, Chantler, and Dai]{lane1998robust}
David~M Lane, Mike~J Chantler, and Dongyong Dai.
\newblock Robust tracking of multiple objects in sector-scan sonar image
  sequences using optical flow motion estimation.
\newblock \emph{IEEE Journal of Oceanic Engineering}, 23\penalty0 (1):\penalty0
  31--46, 1998.

\bibitem[Lei et~al.(2023)Lei, Wei, Zhang, Li, and Zhang]{lei2023medlsam}
Wenhui Lei, Xu Wei, Xiaofan Zhang, Kang Li, and Shaoting Zhang.
\newblock Medlsam: Localize and segment anything model for 3d medical images.
\newblock \emph{arXiv}, 2023.

\bibitem[Li et~al.(2021{\natexlab{a}})Li, Liu, Chen, Bai, Lin, and
  Ling]{li2021hierarchical}
Gongyang Li, Zhi Liu, Minyu Chen, Zhen Bai, Weisi Lin, and Haibin Ling.
\newblock Hierarchical alternate interaction network for rgb-d salient object
  detection.
\newblock \emph{TIP}, 30:\penalty0 3528--3542, 2021{\natexlab{a}}.

\bibitem[Li et~al.(2020)Li, Rigall, Dong, and Chen]{li2020mas3k}
Lin Li, Eric Rigall, Junyu Dong, and Geng Chen.
\newblock Mas3k: An open dataset for marine animal segmentation.
\newblock In \emph{International Symposium on Benchmarking, Measuring and
  Optimization}, pages 194--212. Springer, 2020.

\bibitem[Li et~al.(2021{\natexlab{b}})Li, Dong, Rigall, Zhou, Dong, and
  Chen]{li2021marine}
Lin Li, Bo Dong, Eric Rigall, Tao Zhou, Junyu Dong, and Geng Chen.
\newblock Marine animal segmentation.
\newblock \emph{TCSVT}, 32\penalty0 (4):\penalty0 2303--2314,
  2021{\natexlab{b}}.

\bibitem[Lin et~al.(2017)Lin, Doll{\'a}r, Girshick, He, Hariharan, and
  Belongie]{lin2017feature}
Tsung-Yi Lin, Piotr Doll{\'a}r, Ross Girshick, Kaiming He, Bharath Hariharan,
  and Serge Belongie.
\newblock Feature pyramid networks for object detection.
\newblock In \emph{ICCV}, pages 2117--2125, 2017.

\bibitem[Liu et~al.(2022)Liu, Zhang, and Barnes]{liu2022modeling}
Jiawei Liu, Jing Zhang, and Nick Barnes.
\newblock Modeling aleatoric uncertainty for camouflaged object detection.
\newblock In \emph{WACV}, pages 1445--1454, 2022.

\bibitem[Liu et~al.(2019)Liu, Hou, Cheng, Feng, and Jiang]{liu2019simple}
Jiang-Jiang Liu, Qibin Hou, Ming-Ming Cheng, Jiashi Feng, and Jianmin Jiang.
\newblock A simple pooling-based design for real-time salient object detection.
\newblock In \emph{CVPR}, pages 3917--3926, 2019.

\bibitem[Liu and Yu(2022)]{liu2022underwater}
Lidan Liu and Weiwei Yu.
\newblock Underwater image saliency detection via attention-based mechanism.
\newblock In \emph{Journal of Physics: Conference Series}, page 012012. IOP
  Publishing, 2022.

\bibitem[Liu et~al.(2021{\natexlab{a}})Liu, Zhang, Shao, and
  Han]{liu2021learning}
Nian Liu, Ni Zhang, Ling Shao, and Junwei Han.
\newblock Learning selective mutual attention and contrast for rgb-d saliency
  detection.
\newblock \emph{TPAMI}, 44\penalty0 (12):\penalty0 9026--9042,
  2021{\natexlab{a}}.

\bibitem[Liu et~al.(2021{\natexlab{b}})Liu, Lin, Cao, Hu, Wei, Zhang, Lin, and
  Guo]{liu2021swin}
Ze Liu, Yutong Lin, Yue Cao, Han Hu, Yixuan Wei, Zheng Zhang, Stephen Lin, and
  Baining Guo.
\newblock Swin transformer: Hierarchical vision transformer using shifted
  windows.
\newblock In \emph{ICCV}, pages 10012--10022, 2021{\natexlab{b}}.

\bibitem[Liu et~al.(2021{\natexlab{c}})Liu, Wang, Tu, Xiao, and
  Tang]{liu2021tritransnet}
Zhengyi Liu, Yuan Wang, Zhengzheng Tu, Yun Xiao, and Bin Tang.
\newblock Tritransnet: Rgb-d salient object detection with a triplet
  transformer embedding network.
\newblock In \emph{ACMMM}, pages 4481--4490, 2021{\natexlab{c}}.

\bibitem[Loshchilov and Hutter(2017)]{loshchilov2017decoupled}
Ilya Loshchilov and Frank Hutter.
\newblock Decoupled weight decay regularization.
\newblock \emph{arXiv}, 2017.

\bibitem[Lv et~al.(2021)Lv, Zhang, Dai, Li, Liu, Barnes, and
  Fan]{lv2021simultaneously}
Yunqiu Lv, Jing Zhang, Yuchao Dai, Aixuan Li, Bowen Liu, Nick Barnes, and
  Deng-Ping Fan.
\newblock Simultaneously localize, segment and rank the camouflaged objects.
\newblock In \emph{CVPR}, pages 11591--11601, 2021.

\bibitem[Ma et~al.(2021)Ma, Xia, and Li]{ma2021pyramidal}
Mingcan Ma, Changqun Xia, and Jia Li.
\newblock Pyramidal feature shrinking for salient object detection.
\newblock In \emph{AAAI}, pages 2311--2318, 2021.

\bibitem[Mei et~al.(2021)Mei, Ji, Wei, Yang, Wei, and Fan]{mei2021camouflaged}
Haiyang Mei, Ge-Peng Ji, Ziqi Wei, Xin Yang, Xiaopeng Wei, and Deng-Ping Fan.
\newblock Camouflaged object segmentation with distraction mining.
\newblock In \emph{CVPR}, pages 8772--8781, 2021.

\bibitem[Ng and Henikoff(2003)]{ng2003sift}
Pauline~C Ng and Steven Henikoff.
\newblock Sift: Predicting amino acid changes that affect protein function.
\newblock \emph{NAS}, 31\penalty0 (13):\penalty0 3812--3814, 2003.

\bibitem[Pang et~al.(2020)Pang, Zhao, Zhang, and Lu]{pang2020multi}
Youwei Pang, Xiaoqi Zhao, Lihe Zhang, and Huchuan Lu.
\newblock Multi-scale interactive network for salient object detection.
\newblock In \emph{CVPR}, pages 9413--9422, 2020.

\bibitem[Pang et~al.(2022)Pang, Zhao, Xiang, Zhang, and Lu]{pang2022zoom}
Youwei Pang, Xiaoqi Zhao, Tian-Zhu Xiang, Lihe Zhang, and Huchuan Lu.
\newblock Zoom in and out: A mixed-scale triplet network for camouflaged object
  detection.
\newblock In \emph{CVPR}, pages 2160--2170, 2022.

\bibitem[Piao et~al.(2019)Piao, Ji, Li, Zhang, and Lu]{piao2019depth}
Yongri Piao, Wei Ji, Jingjing Li, Miao Zhang, and Huchuan Lu.
\newblock Depth-induced multi-scale recurrent attention network for saliency
  detection.
\newblock In \emph{ICCV}, pages 7254--7263, 2019.

\bibitem[Piao et~al.(2021)Piao, Wang, Zhang, and Lu]{piao2021mfnet}
Yongri Piao, Jian Wang, Miao Zhang, and Huchuan Lu.
\newblock Mfnet: Multi-filter directive network for weakly supervised salient
  object detection.
\newblock In \emph{ICCV}, pages 4136--4145, 2021.

\bibitem[Priyadarshni and Kolekar(2020)]{priyadarshni2020underwater}
Divya Priyadarshni and MaheshKumar~H Kolekar.
\newblock Underwater object detection and tracking.
\newblock In \emph{Soft Computing}, pages 837--846. Springer, 2020.

\bibitem[Priyadharsini and Sharmila(2019)]{priyadharsini2019object}
R Priyadharsini and T~Sree Sharmila.
\newblock Object detection in underwater acoustic images using edge based
  segmentation method.
\newblock \emph{Procedia Computer Science}, 165:\penalty0 759--765, 2019.

\bibitem[Qin et~al.(2019)Qin, Zhang, Huang, Gao, Dehghan, and
  Jagersand]{qin2019basnet}
Xuebin Qin, Zichen Zhang, Chenyang Huang, Chao Gao, Masood Dehghan, and Martin
  Jagersand.
\newblock Basnet: Boundary-aware salient object detection.
\newblock In \emph{CVPR}, pages 7479--7489, 2019.

\bibitem[Qin et~al.(2020)Qin, Zhang, Huang, Dehghan, Zaiane, and
  Jagersand]{qin2020u2}
Xuebin Qin, Zichen Zhang, Chenyang Huang, Masood Dehghan, Osmar~R Zaiane, and
  Martin Jagersand.
\newblock U2-net: Going deeper with nested u-structure for salient object
  detection.
\newblock \emph{PR}, 106:\penalty0 107404, 2020.

\bibitem[Qu et~al.(2017)Qu, He, Zhang, Tian, Tang, and Yang]{qu2017rgbd}
Liangqiong Qu, Shengfeng He, Jiawei Zhang, Jiandong Tian, Yandong Tang, and
  Qingxiong Yang.
\newblock Rgbd salient object detection via deep fusion.
\newblock \emph{TIP}, 26\penalty0 (5):\penalty0 2274--2285, 2017.

\bibitem[Ranftl et~al.(2021)Ranftl, Bochkovskiy, and Koltun]{ranftl2021vision}
Ren{\'e} Ranftl, Alexey Bochkovskiy, and Vladlen Koltun.
\newblock Vision transformers for dense prediction.
\newblock In \emph{ICCV}, pages 12179--12188, 2021.

\bibitem[Shan and Zhang(2023)]{shan2023robustness}
Xinru Shan and Chaoning Zhang.
\newblock Robustness of segment anything model (sam) for autonomous driving in
  adverse weather conditions.
\newblock \emph{arXiv}, 2023.

\bibitem[Shihavuddin et~al.(2013)Shihavuddin, Gracias, Garcia, Escartin, and
  Pedersen]{shihavuddin2013automated}
ASM Shihavuddin, Nuno Gracias, Rafael Garcia, Javier Escartin, and Rolf~Birger
  Pedersen.
\newblock Automated classification and thematic mapping of bacterial mats in
  the north sea.
\newblock In \emph{OCEANS}, pages 1--8. IEEE, 2013.

\bibitem[Sun et~al.(2021)Sun, Chen, Zhou, Zhang, and Liu]{sun2021context}
Yujia Sun, Geng Chen, Tao Zhou, Yi Zhang, and Nian Liu.
\newblock Context-aware cross-level fusion network for camouflaged object
  detection.
\newblock \emph{arXiv}, 2021.

\bibitem[Wang et~al.(2020)Wang, Chen, Zhou, Zhang, Jin, and
  Gai]{wang2020progressive}
Bo Wang, Quan Chen, Min Zhou, Zhiqiang Zhang, Xiaogang Jin, and Kun Gai.
\newblock Progressive feature polishing network for salient object detection.
\newblock In \emph{AAAI}, pages 12128--12135, 2020.

\bibitem[Wang et~al.(2004)Wang, Bovik, Sheikh, and Simoncelli]{wang2004image}
Zhou Wang, Alan~C Bovik, Hamid~R Sheikh, and Eero~P Simoncelli.
\newblock Image quality assessment: from error visibility to structural
  similarity.
\newblock \emph{IEEE TIP}, 13\penalty0 (4):\penalty0 600--612, 2004.

\bibitem[Wei et~al.(2020{\natexlab{a}})Wei, Wang, and Huang]{wei2020f3net}
Jun Wei, Shuhui Wang, and Qingming Huang.
\newblock F$^3$net: fusion, feedback and focus for salient object detection.
\newblock In \emph{AAAI}, pages 12321--12328, 2020{\natexlab{a}}.

\bibitem[Wei et~al.(2020{\natexlab{b}})Wei, Wang, Wu, Su, Huang, and
  Tian]{wei2020label}
Jun Wei, Shuhui Wang, Zhe Wu, Chi Su, Qingming Huang, and Qi Tian.
\newblock Label decoupling framework for salient object detection.
\newblock In \emph{CVPR}, pages 13025--13034, 2020{\natexlab{b}}.

\bibitem[Wu et~al.(2019{\natexlab{a}})Wu, Su, and Huang]{wu2019cascaded}
Zhe Wu, Li Su, and Qingming Huang.
\newblock Cascaded partial decoder for fast and accurate salient object
  detection.
\newblock In \emph{CVPR}, pages 3907--3916, 2019{\natexlab{a}}.

\bibitem[Wu et~al.(2019{\natexlab{b}})Wu, Su, and Huang]{wu2019stacked}
Zhe Wu, Li Su, and Qingming Huang.
\newblock Stacked cross refinement network for edge-aware salient object
  detection.
\newblock In \emph{ICCV}, pages 7264--7273, 2019{\natexlab{b}}.

\bibitem[Xu et~al.(2021)Xu, Liang, Liang, and Chen]{xu2021locate}
Binwei Xu, Haoran Liang, Ronghua Liang, and Peng Chen.
\newblock Locate globally, segment locally: A progressive architecture with
  knowledge review network for salient object detection.
\newblock In \emph{AAAI}, pages 3004--3012, 2021.

\bibitem[Xu et~al.(2023)Xu, Su, and Liu]{xu2023aquasam}
Muduo Xu, Jianhao Su, and Yutao Liu.
\newblock Aquasam: Underwater image foreground segmentation.
\newblock \emph{arXiv}, 2023.

\bibitem[Yan et~al.(2022)Yan, Wan, and Zhang]{yan2022fully}
Tianyu Yan, Zifu Wan, and Pingping Zhang.
\newblock Fully transformer network for change detection of remote sensing
  images.
\newblock In \emph{ACCV}, pages 1691--1708, 2022.

\bibitem[Yan et~al.(2023)Yan, Wan, Zhang, Cheng, and Lu]{10292873}
Tianyu Yan, Zifu Wan, Pingping Zhang, Gong Cheng, and Huchuan Lu.
\newblock Transy-net: Learning fully transformer networks for change detection
  of remote sensing images.
\newblock \emph{TGRS}, 61:\penalty0 1--12, 2023.

\bibitem[Yang et~al.(2021)Yang, Lin, Lin, Jiang, and Liu]{yang2021progressive}
Sheng Yang, Weisi Lin, Guosheng Lin, Qiuping Jiang, and Zichuan Liu.
\newblock Progressive self-guided loss for salient object detection.
\newblock \emph{TIP}, 30:\penalty0 8426--8438, 2021.

\bibitem[Yuan et~al.(2017)Yuan, Li, Kim, Cai, and Feng]{yuan2017reversion}
Yuchen Yuan, Changyang Li, Jinman Kim, Weidong Cai, and David~Dagan Feng.
\newblock Reversion correction and regularized random walk ranking for saliency
  detection.
\newblock \emph{TIP}, 27\penalty0 (3):\penalty0 1311--1322, 2017.

\bibitem[Zhang et~al.(2021{\natexlab{a}})Zhang, Cong, Lin, Ma, Li, Zhao, and
  Kwong]{zhang2021cross}
Chen Zhang, Runmin Cong, Qinwei Lin, Lin Ma, Feng Li, Yao Zhao, and Sam Kwong.
\newblock Cross-modality discrepant interaction network for rgb-d salient
  object detection.
\newblock In \emph{ACMMM}, pages 2094--2102, 2021{\natexlab{a}}.

\bibitem[Zhang et~al.(2020{\natexlab{a}})Zhang, Tian, and Han]{zhang2020few}
Dingwen Zhang, Haibin Tian, and Jungong Han.
\newblock Few-cost salient object detection with adversarial-paced learning.
\newblock \emph{ANIPS}, 33:\penalty0 12236--12247, 2020{\natexlab{a}}.

\bibitem[Zhang et~al.(2023{\natexlab{a}})Zhang, Liang, Yang, Zou, Ye, Liu, and
  Bai]{zhang2023sam3d}
Dingyuan Zhang, Dingkang Liang, Hongcheng Yang, Zhikang Zou, Xiaoqing Ye, Zhe
  Liu, and Xiang Bai.
\newblock Sam3d: Zero-shot 3d object detection via segment anything model.
\newblock \emph{arXiv}, 2023{\natexlab{a}}.

\bibitem[Zhang et~al.(2020{\natexlab{b}})Zhang, Fan, Dai, Anwar, Saleh, Zhang,
  and Barnes]{zhang2020uc}
Jing Zhang, Deng-Ping Fan, Yuchao Dai, Saeed Anwar, Fatemeh~Sadat Saleh, Tong
  Zhang, and Nick Barnes.
\newblock Uc-net: Uncertainty inspired rgb-d saliency detection via conditional
  variational autoencoders.
\newblock In \emph{CVPR}, pages 8582--8591, 2020{\natexlab{b}}.

\bibitem[Zhang and Liu(2023)]{zhang2023customized}
Kaidong Zhang and Dong Liu.
\newblock Customized segment anything model for medical image segmentation.
\newblock \emph{arXiv}, 2023.

\bibitem[Zhang et~al.(2023{\natexlab{b}})Zhang, Liu, Zhang, Wu, Yu, Holmes,
  Feng, Dai, Li, Li, et~al.]{zhang2023segment}
Lian Zhang, Zhengliang Liu, Lu Zhang, Zihao Wu, Xiaowei Yu, Jason Holmes,
  Hongying Feng, Haixing Dai, Xiang Li, Quanzheng Li, et~al.
\newblock Segment anything model (sam) for radiation oncology.
\newblock \emph{arXiv}, 2023{\natexlab{b}}.

\bibitem[Zhang et~al.(2021{\natexlab{b}})Zhang, Jiang, Fu, and
  Zhao]{zhang2021bts}
Wenbo Zhang, Yao Jiang, Keren Fu, and Qijun Zhao.
\newblock Bts-net: Bi-directional transfer-and-selection network for rgb-d
  salient object detection.
\newblock In \emph{ICME}, pages 1--6. IEEE, 2021{\natexlab{b}}.

\bibitem[Zhao et~al.(2020{\natexlab{a}})Zhao, Zhao, Li, and
  Chen]{zhao2020depth}
Jiawei Zhao, Yifan Zhao, Jia Li, and Xiaowu Chen.
\newblock Is depth really necessary for salient object detection?
\newblock In \emph{ACMMM}, pages 1745--1754, 2020{\natexlab{a}}.

\bibitem[Zhao et~al.(2019)Zhao, Liu, Fan, Cao, Yang, and Cheng]{zhao2019egnet}
Jia-Xing Zhao, Jiang-Jiang Liu, Deng-Ping Fan, Yang Cao, Jufeng Yang, and
  Ming-Ming Cheng.
\newblock Egnet: Edge guidance network for salient object detection.
\newblock In \emph{ICCV}, pages 8779--8788, 2019.

\bibitem[Zhao et~al.(2023)Zhao, Zhang, Tang, Gu, and Zhu]{zhao2023enlighten}
Qihan Zhao, Xiaofeng Zhang, Hao Tang, Chaochen Gu, and Shanying Zhu.
\newblock Enlighten-anything: When segment anything model meets low-light image
  enhancement.
\newblock \emph{arXiv}, 2023.

\bibitem[Zhao and Wu(2019)]{zhao2019pyramid}
Ting Zhao and Xiangqian Wu.
\newblock Pyramid feature attention network for saliency detection.
\newblock In \emph{CVPR}, pages 3085--3094, 2019.

\bibitem[Zhao et~al.(2020{\natexlab{b}})Zhao, Zhang, Pang, Lu, and
  Zhang]{zhao2020single}
Xiaoqi Zhao, Lihe Zhang, Youwei Pang, Huchuan Lu, and Lei Zhang.
\newblock A single stream network for robust and real-time rgb-d salient object
  detection.
\newblock In \emph{ECCV}, pages 646--662. Springer, 2020{\natexlab{b}}.

\bibitem[Zhao et~al.(2021)Zhao, Xia, Xie, and Li]{zhao2021complementary}
Zhirui Zhao, Changqun Xia, Chenxi Xie, and Jia Li.
\newblock Complementary trilateral decoder for fast and accurate salient object
  detection.
\newblock In \emph{ACMMM}, pages 4967--4975, 2021.

\bibitem[Zheng et~al.(2021)Zheng, Lu, Zhao, Zhu, Luo, Wang, Fu, Feng, Xiang,
  Torr, et~al.]{zheng2021rethinking}
Sixiao Zheng, Jiachen Lu, Hengshuang Zhao, Xiatian Zhu, Zekun Luo, Yabiao Wang,
  Yanwei Fu, Jianfeng Feng, Tao Xiang, Philip~HS Torr, et~al.
\newblock Rethinking semantic segmentation from a sequence-to-sequence
  perspective with transformers.
\newblock In \emph{CVPR}, pages 6881--6890, 2021.

\bibitem[Zhou et~al.(2021)Zhou, Fu, Chen, Zhou, Fan, and
  Shao]{zhou2021specificity}
Tao Zhou, Huazhu Fu, Geng Chen, Yi Zhou, Deng-Ping Fan, and Ling Shao.
\newblock Specificity-preserving rgb-d saliency detection.
\newblock In \emph{ICCV}, pages 4681--4691, 2021.

\bibitem[Zhou et~al.(2018)Zhou, Rahman~Siddiquee, Tajbakhsh, and
  Liang]{zhou2018unet++}
Zongwei Zhou, Md~Mahfuzur Rahman~Siddiquee, Nima Tajbakhsh, and Jianming Liang.
\newblock Unet++: A nested u-net architecture for medical image segmentation.
\newblock In \emph{MICCAI}, pages 3--11. Springer, 2018.

\end{thebibliography}
}

\end{document}